\documentclass[journal, comsoc]{IEEEtran}

\usepackage{cite}
\usepackage{amsmath,amssymb,amsfonts}
\usepackage{graphicx}
\usepackage{textcomp}
\usepackage{xcolor}
\definecolor{GRAY}{RGB}{236,236,236}
\definecolor{DARKGRAY}{RGB}{230,230,230}
\usepackage{colortbl}

\usepackage{subfigure}
\usepackage{xcolor}
\usepackage{comment}
\usepackage{caption}
\usepackage{multirow}
\usepackage{url}
\usepackage{enumerate}
\usepackage{color}
\usepackage{footnote}
\usepackage{pifont}
\usepackage{ulem}
\usepackage{booktabs}
\usepackage{wasysym}
\usepackage{algpseudocode}
\usepackage[ruled,vlined,linesnumbered]{algorithm2e}
\let\oldnl\nl
\newcommand{\nonl}{\renewcommand{\nl}{\let\nl\oldnl}}

\usepackage{amsmath}

\usepackage{hyperref}
\hypersetup{
     colorlinks = true,
     linkcolor = blue,
     anchorcolor = red,
     citecolor = magenta,
     filecolor = blue,
     urlcolor = black,
}

\usepackage{threeparttable}
\usepackage{makecell}

\begin{document}

\title{\textsc{IronForge}: An Open, Secure, Fair, Decentralized Federated Learning}

\author{\IEEEauthorblockN{
Guangsheng Yu\IEEEauthorrefmark{1}, 
Xu Wang\IEEEauthorrefmark{2}, 
Caijun Sun\IEEEauthorrefmark{4}, 
Qin Wang\IEEEauthorrefmark{1}, 
Ping Yu\IEEEauthorrefmark{3}, \\
Wei Ni\IEEEauthorrefmark{1},
Renping Liu\IEEEauthorrefmark{2},
Xiwei Xu\IEEEauthorrefmark{1}
}\\
\IEEEauthorrefmark{1}\textit{CSIRO Data61, Australia}\\
\IEEEauthorrefmark{2}\textit{University of Technology Sydney,  Australia} \\
\IEEEauthorrefmark{3}\textit{Harbin University of Technology, China}\\
\IEEEauthorrefmark{4}\textit{Zhejiang Lab, China} 
}

\maketitle

\thispagestyle{plain}
\pagestyle{plain}

\begin{abstract}
Federated learning (FL) provides an effective machine learning (ML) architecture to protect data privacy in a distributed manner. However, the inevitable network asynchrony, the over-dependence on a central coordinator, and the lack of an open and fair incentive mechanism collectively hinder its further development. We propose \textsc{IronForge}, a new generation of FL framework, that features a Directed Acyclic Graph (DAG)-based data structure and eliminates the need for central coordinators to achieve fully decentralized operations. \textsc{IronForge} runs in a public and open network, and launches a fair incentive mechanism by enabling state consistency in the DAG, so that the system fits in networks where training resources are unevenly distributed. In addition, dedicated defense strategies against prevalent FL attacks on incentive fairness and data privacy are presented to ensure the security of \textsc{IronForge}. Experimental results based on a newly developed testbed FLSim 
highlight the superiority of \textsc{IronForge} to the existing prevalent FL frameworks under various specifications in performance, fairness, and security. To the best of our knowledge, \textsc{IronForge} is the first secure and fully decentralized FL framework that can be applied in open networks with realistic network and training settings.

\end{abstract}

\renewcommand{\arraystretch}{1.3}

\section{Introduction}\label{sec_intro}
Federated learning (FL), officially introduced by Google in 2017~\cite{fl-2}, has become the preference to aggregate data from distributed ends without breaching data privacy~\cite{fl-1,fl-2}. By aggregating huge data with comprehensive extracted features in FL, critical issues such as model overfitting can be significantly addressed~\cite{fl-3}. However, \ding{192} the inevitable network asynchrony, \ding{193} the over-dependence on a central coordinator, and \ding{194} the lack of an open and fair incentive mechanism hinder the further development of FL in large and open scenarios~\cite{9084352}.

Traditional FL considers no or low delay throughout an aggregation process, namely, synchronous FL. However, network synchrony is unrealistic due to the inevitable capacity limit of computation, bandwidth, and storage, as well as the imbalanced capacities among the distributed participants. Thus, recent studies propose pseudo-asynchronous FL~\cite{9093123} and asynchronous FL~\cite{async-fl}. The aggregation of pseudo-asynchronous FL allows a short interval for collecting the model caches in order to ensure that the number of models aggregated can be sufficiently large, while 
the central coordinator immediately updates the global model once receiving a new local model from any idle participants in asynchronous FL.

Neither pseudo-asynchronous FL nor asynchronous FL can tolerate the single-point-of-failure (SPoF) of the central coordinator or even a malicious and corrupted coordinator (issue-\ding{193}). The over-dependence on the central coordinator could potentially degrade the system availability and the training flexibility in the sense that an FL network may be confined to specific training domains or tasks determined by the coordinator. Participants in many existing studies~\cite{8994206,9170265,9524833}, once opting in an FL network, would have to obey the defined training target with no flexibility to go for different tasks at will.

In addition to the weak training flexibility, the lack of an open and fair incentive mechanism results in participants who have fewer resources and a weaker capacity not willing to contribute their resources to the global aggregation. This issue deteriorates particularly in FL networks where resources are not evenly distributed, and potentially leads to the model overfitting and weak generality against contingencies. Although 
the authors of~\cite{fl-incentive} survey the incentive mechanisms in FL, all mentioned frameworks require a central coordinator, also leading to issue-\ding{193}.

Existing studies propose to replace the central coordinator with a committee running a consensus process in a blockchain network to prevent the SPoF or a corrupted coordinator. Meanwhile, by sharing the model collection during the consensus in the committee, pseudo-asynchronous FL can be achieved in a decentralized manner, i.e., BlockFL~\cite{8994206, bc-fl-1,bc-fl-2, sharding-fl, 8647616}. Considering only issue-\ding{193} being solved and issue-\ding{192} being partially solved by BlockFL, the authors of ~\cite{9524833} introduce a Directed Acyclic Graph (DAG)-based FL where both issue-\ding{192} and issue-\ding{193} are solved using the concept of asynchronous FL~\cite{async-fl} to fully decentralize the FL process. However, the paper~\cite{9524833} only considers an ideal network in which the training resources are evenly distributed. Moreover, the approach to enabling state consistency for a secure and fair incentive mechanism (issue-\ding{194}) is missing in~\cite{9524833}, which results in difficulty in adopting the mechanism in a public and open network.

\begin{table*}[!ht]
    \centering
    \caption{Qualitative comparisons between the proposed $\textsc{IronForge}$ and the existing FL frameworks}\label{tab:fl_comparison}

    \begin{threeparttable}
    \resizebox{\textwidth}{!}{
    \begin{tabular}{>{\columncolor{DARKGRAY}}lllllll}
    \toprule
     \rowcolor{DARKGRAY}\multicolumn{1}{l}{\textbf{FL Framework}}   &
    \multicolumn{1}{l}{\textbf{Data Structure}}   & \multicolumn{1}{l}{\textbf{Data Asynchrony}} & 
    \multicolumn{1}{l}{\textbf{Decentralization}} &
    \multicolumn{1}{l}{\textbf{Openness}} & \multicolumn{1}{l}{\textbf{Incentive}}  &
    \multicolumn{1}{l}{\textbf{Security}}
    \\
    \midrule
    
  Google FL~\cite{fl-1} & Isolated models & Synchronous &  Centralized  & Private & \Circle &\Circle\\
  Asynchronous FL~\cite{async-fl} & Isolated models & Asynchronous &  Centralized  & Private & \Circle & \Circle  \\
  Block FL~\cite{8733825} & Blockchain & Synchronous &  Decentralized  & Private & Reward & \Circle  \\
  DAG FL~\cite{9524833} & DAG & Asynchronous &  Decentralized  & Public & Reward & Poisoning/Backdoor/Lazy  \\
  \textsc{IronForge} & DAG & Asynchronous &  Decentralized  & Public & Reward, Penalty& Poisoning/Backdoor/\textbf{Stealing}\tnote{*} /\textbf{Collusion}     \\

   \bottomrule
    \end{tabular}

    }
    \begin{tablenotes}
        \scriptsize
        \item[*] The stealing attack considered in this paper includes the traditional lazy attack. \\
        The difference is that stealing attackers not only upload their previous models, but also fake the ownership of others' previous models.
        \item[\Circle] Lack of corresponding designs.
    \end{tablenotes}
    \end{threeparttable}
\end{table*}

We propose \textsc{IronForge} that is an open, secure, fair, and decentralized FL system. \textsc{IronForge} solves the above mentioned pain points at one time. \textbf{Openness: }It features a DAG-based data structure in an open network. \textbf{Decentralization: }The need for a central coordinator is eliminated throughout the process by \textsc{IronForge}, inheriting from the concept of asynchronous FL. As a result, the models are maintained in a decentralized manner by all participants. 
\textbf{Fairness: }\textsc{IronForge} considers a practical scenario, where resources are unevenly distributed among users. Each user, based on its resource amount, selects several existing models, verifies the correctness and evaluates the model accuracy over the local dataset, and conducts the aggregation. 
\textsc{IronForge} also enables state consistency, by using which an open and fair incentive mechanism can be established to motivate more participants. \textbf{Security: }Moreover, dedicated defense strategies against malicious attacks on incentive fairness, and against dataset privacy breaching are presented to ensure the security of \textsc{IronForge}. The key contributions are as follows.

\begin{itemize}
\item[$\triangleright$] We propose a fully decentralized FL framework, namely, \textsc{IronForge}, which features a DAG-based data structure. \textsc{IronForge} addresses the network asynchrony typically undergone in an FL process, and improves the motivation of agents participating in the process in an open environment by enabling reliable token rewards with strong consistency and model prediction accuracy.

\item[$\triangleright$] We specifically design a new validation mechanism guarding against well-known FL attacks, including model poisoning attacks, backdoor attacks, lazy attacks, and model stealing attacks, among which the model of stealing attack has never been considered in any existing FL frameworks. By making use of noise-enabled Proof-of-Learning (PoL) to validate the gradient descent process, any malicious behaviors, such as faking the ownership or directly using the existing models, or embezzling the rewards for their conspirator by claiming a falsified source list, can be captured and given punishments. 

\item[$\triangleright$] We build a flexible and efficient testbed, named FLSim, to simulate the workflow across all considered FL frameworks in this paper, including the proposed \textsc{IronForge}. We conduct comprehensive experiments based on FLSim, comparing the system performance, security, and fairness between the existing FL frameworks and \textsc{IronForge}. Insights are shed to provide guidelines on how to select strategies in \textsc{IronForge} to meet different requirements.
\end{itemize}

Extensive experiments corroborate that \textsc{IronForge} outperforms the prevalent FL frameworks with and without attacks leveraged, which highlights the holistic solution to the network asynchrony (issue-\ding{192}) and the over-dependence on the central coordinators (issue-\ding{193}). Strictly and approximately monotonic increases of rewards are observed in experiments with increasing CPU cores, memory capacity, and bandwidth in different incentive settings. This indicates that fairness (issue-\ding{194}) can be ensured in \textsc{IronForge} under various definitions of fairness.

The rest of the paper is organized as follows.
Section~\ref{sec_intro} gives the introduction, followed by related works in~Section~\ref{sec_relatedwork}.
Section~\ref {sec_design} provides the system overview and Section~\ref{sec_DAG} details the design of \textsc{IronForge}. Section~\ref{sec_develop} presents our implementation based on a new testbed with comprehensive experimental results. Section \ref{sec_analysis} discusses system security and properties. Finally, Section~\ref{sec_con} concludes this work.

\section{Related Work}
\label{sec_relatedwork}


A conventional synchronous FL framework is constructed by a central coordinator and numbers of nodes, which maintains the global model and perform FL iterations, respectively~\cite{fl-1}. The coordinator periodically distributes the latest global model to the nodes, and then the nodes independently train the model with their local data and upload the trained local models to the coordinator \cite{8889996}. After receiving updated models from nodes, the coordinator aggregates all the local models as a new global model. 
Such synchronous FL framework can hardly be adapted to large-scale and heterogeneous networks, where asynchrony is non-negligible.

The issue of data asynchrony is tackled by the asynchronous FL enabling nodes to train the global model from central coordinators at any time, and the coordinators can update the global model immediately when any local model is collected.
In~\cite{8647616}, the authors introduced a cache layer between the coordinator and local nodes. Each node trains the global model with its local data and uploads its model to the cache. The coordinator periodically aggregates the local model in the cache and generates a new global model. 
Semi-asynchronous FL protocols address the problems in FL such as low round efficiency and poor convergence rate happened in asynchronous FL. The system~\cite{9093123} incorporates a client selection algorithm decoupling the coordinator and the selected clients for a reduction of average round time.
The authors of~\cite{tian2021towards} proposed an asynchronous federating-based detection approach for end devices. A pre-shared data training strategy for non-independent-and-identically-distributed (non-IID) data is developed to avoid convergence divergence under the non-IID patterns. After the collaborative model training procedure, each client further conducts an additional local training process to fit respective patterns.

The aforementioned FL frameworks require central coordinators to schedule model training and aggregate models. The centralized architecture suffers inherent security risks, such as SPoF and malicious central coordinator, and limited scalability with the bottleneck of the central coordinator.
The most recent Distributed Ledger Technology (DLT) holds the potential to decentralize FL systems~\cite{WANG201910,sharding}. Two key technologies in DLT are blockchain and DAG.
In blockchain, a group of miners run the consensus protocol to generate hash-chained data blocks, which are assembled from transactional data, and synchronize the chained blocks. Blockchain assures strong consistency among blockchain nodes and enables smart contracts to be executed across the blockchain network in a consistent and trustworthy way. In DAG, transactions from decentralized DAG users are organized in a DAG structure where directed edges indicate the reference relationship between the transactions. DAG can achieve high throughput with short latency compared with blockchain~\cite{CAO2020480}.

DLT has been developed to remove the central coordinator and decentralize FL networks~\cite{8733825,8998397,9524833}. 
In BlockFL~\cite{8733825, 9833437, 9403374}, decentralized blockchain miners conduct model verification and aggregation. To be specific, miners obtain trained local models from working nodes and other miners. After verification, miners aggregate local models for the updated global models and conduct Proof-of-Work (PoW) to create valid blocks containing the new global models. Then, the blocks are propagated to all miners to start the next FL iteration. The BlockFL relies on the resource-intensive PoW consensus protocol to slow down the system and keep miners synchronized. To reduce overhead and improve scalability, DAG technology \cite{wang2022sok} is introduced to FL networks~\cite{9524833,8998397}, where trained models are updated to a DAG topology by working nodes without any coordination. Working nodes can learn the latest local models in the DAG by exchanging data with other nodes. By themselves, working nodes select and verify aggregate local models and train the models using local datasets. Next, working nodes publish their trained models to the DAG with directed edges indicating the model reference.

Existing works only consider homogeneous networks where the training resources are evenly distributed and thus lack open and fair incentive mechanisms. \textsc{IronForge} proposed by this paper, on the other hand, improves the motivation of participants with rewards for training contributions and penalties for dishonest behaviors.
\textsc{IronForge} also tackles new vulnerabilities in open FL networks, including model stealing attacks where attackers steal models from others and claim rewards from the plagiarized model, and collusion attacks where attackers claim trained models are from conspirators.

\section{System Overview}\label{sec_design}

In this section, we describe \textsc{IronForge} from the aspects of its architecture, workflow, and system assumptions.

\begin{figure}
    \centering
    \includegraphics[width=1\linewidth]{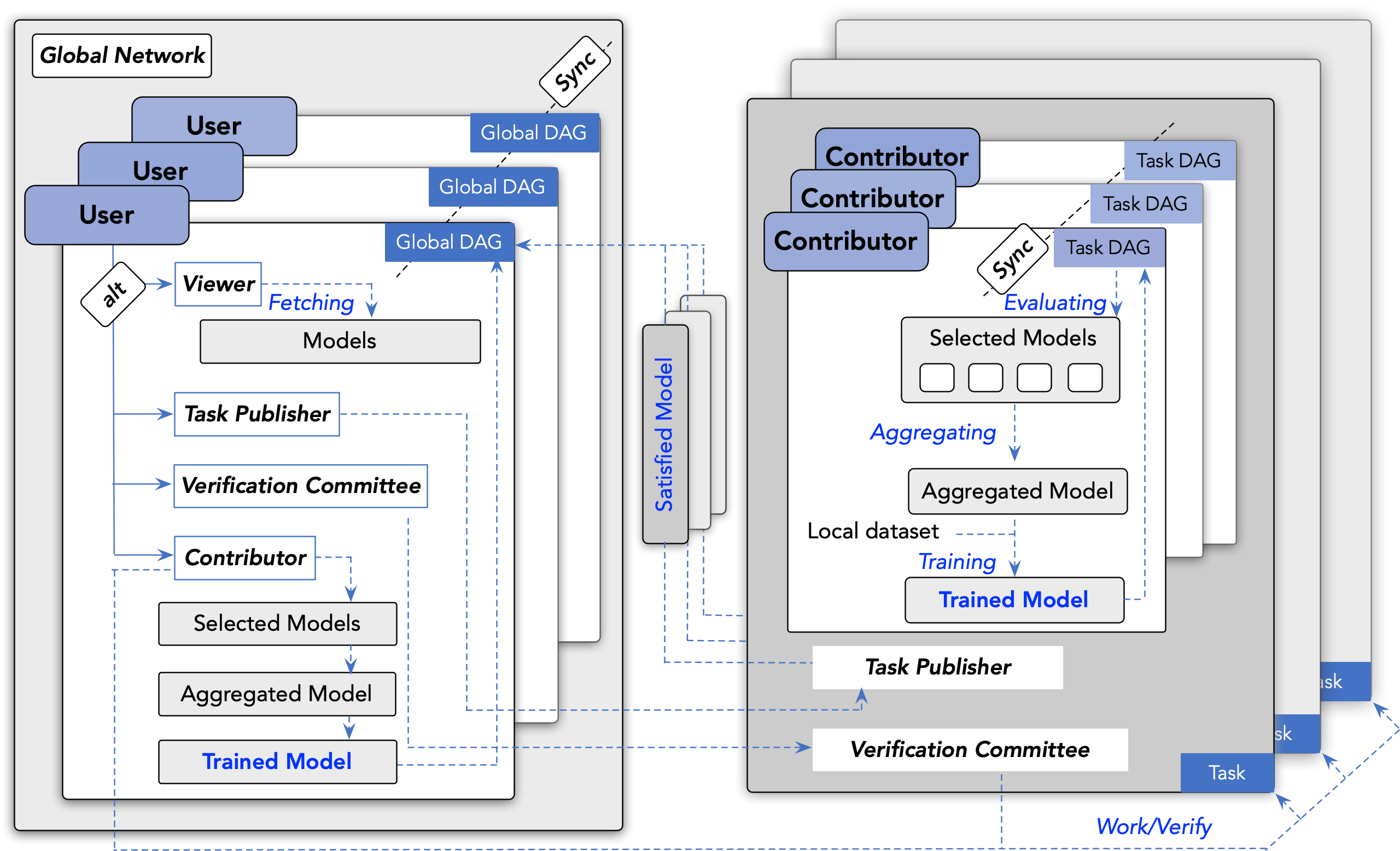}
    \caption{System model of \textsc{IronForge}}
    \label{fig: model}
\end{figure}

\begin{figure*}
    \centering
    \includegraphics[width=1\textwidth]{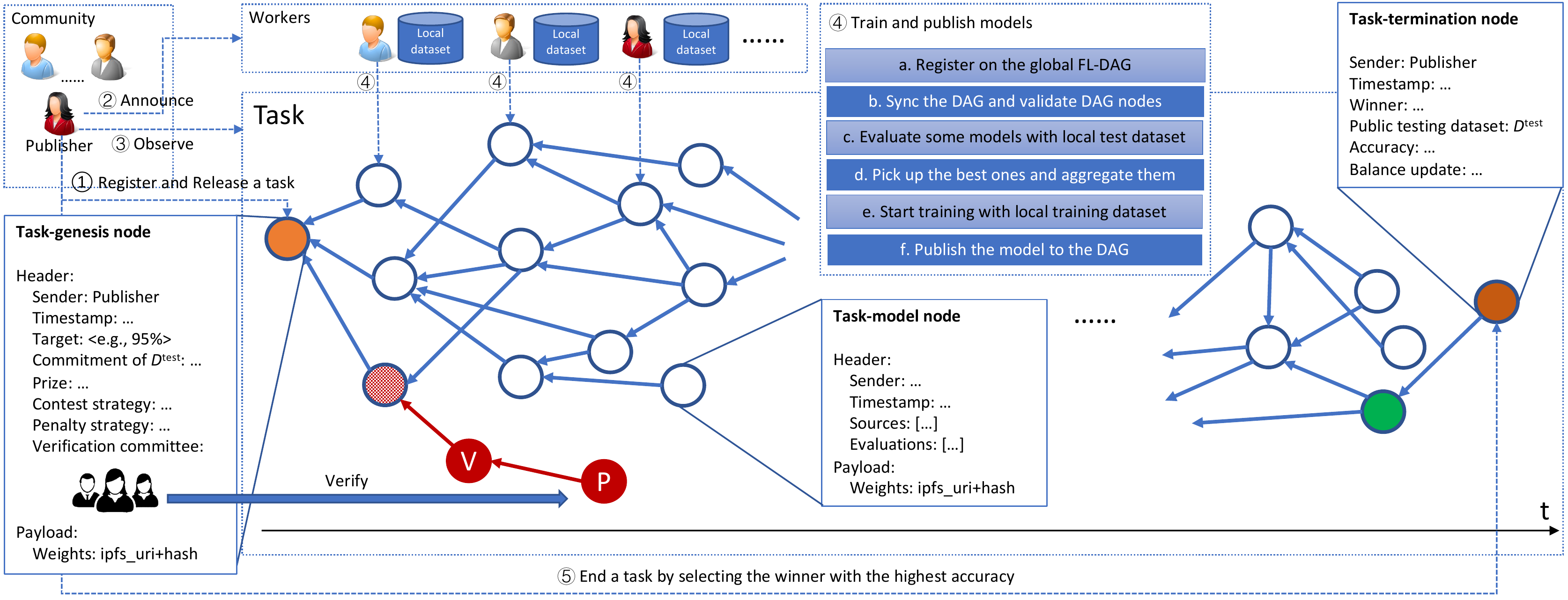}
    \caption{\textbf{\textit{Task-DAG.}} The figure illustrates an overview of starting a new DAG-based FL task, also known as \textit{Task-DAG}. One with aims to improve his model accuracy to a certain target with the help of the community can release a task as the task publisher. Some amount of token is deposited as the prize which will be subsequently awarded to all eligible participants until the winning model is found and selected by the task publisher. The balance update of each participant is recorded in a task-termination node published by the task publisher, and can be subsequently settled by the \textit{Global-DAG} network.}
    \label{fig: task-dag}
\end{figure*}

\begin{figure*}
    \centering
    \includegraphics[width=1\textwidth]{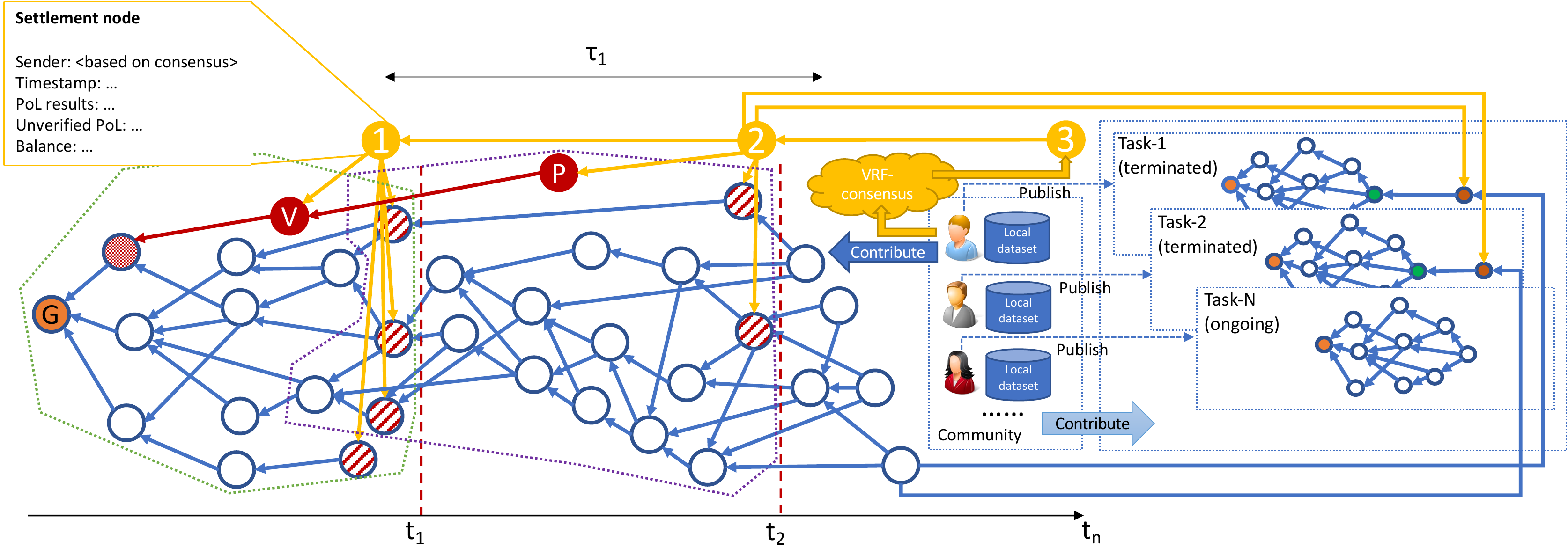}
    \caption{\textbf{\textit{Global-DAG.}} This figure illustrates an overview of the \textit{Global-DAG} network. With the absence of a centralized coordinator, each participant trains a model by selecting and aggregating as many models (including the outcomes of terminated tasks) published by others as possible (based on the local capability). Model targets are not unique and according to different needs, the network can be treated as a global resource pool containing a variety of models. Ones can either find a model which satisfies his local testing dataset from the pool, or make contributions to the pool and obtain token rewards by improving existing models. Token balances are periodically settled (endorsed by a verifiable random function (VRF)-driven consensus) by settlement nodes that employ a chain structure to achieve strong consistency.}
    \label{fig: global-dag}
\end{figure*}

\subsection{System Overview}

We first introduce the roles that participate in the system and present our high-level design.

\smallskip
\noindent\textbf{Architecture.} \textsc{IronForge} is a decentralized FL system that features a DAG-based network structure to tackle the inconsistency in the decentralized FL process, excessive reliance on central coordination, and ineffective motivation of contributing the learning resources at the same time. Specifically, \textsc{IronForge} builds a hybrid architecture (cf. Fig.~\ref{fig: model}) that involves two types of 
DAG, namely, \textit{Task-DAG} and \textit{Global-DAG} (details refer to Fig.~\ref{fig: task-dag} and Fig.~\ref{fig: global-dag}, respectively). 
The training processes in both \textit{Task-DAG} and \textit{Global-DAG} are traceable owing to the DAG data structure. A DAG node  published by a participant consists of a model update and the directed edges of the node indicate the aggregating relationship with existing models during the update, hence no central coordinator required to conduct the training processes. 

\textit{Global-DAG} 
contains a variety of models adopted by all participants, which
can be viewed as a ``unique'' and public model resource pool.
No consistent testing dataset is given in \textit{Global-DAG}. Each user comes to \textit{Global-DAG} and hunts for models that uniquely meet its own local testing dataset.
Without central coordinators, any user can fetch models from the pool for direct uses, release his task requests, or make contributions, such as training on \textit{Global-DAG} or on uncompleted training tasks, or verifying the tasks. 

Each training task is managed by a \textit{Task-DAG}, while \textsc{IronForge} can contain multiple \textit{Task-DAG}s at the same time to handle a range of different training tasks (see the right-hand side in Fig.~\ref{fig: model}).
\textit{Task-DAG}s are task-specific and are released by users who aim at improving their local model prediction accuracy by virtue of the computational powers and resources of others. Within a task, the \textit{Task-DAG} network contains multiple contributors who have the same training target provided by the publisher. The trained models for each task are broadcast and stored in the corresponding \textit{Task-DAG}, and await the check and verification. The satisfied model of a task, observed by the publisher, is subsequently merged into \textit{Global-DAG}, increasing the exposure to the public users. As a result, parallel learning on our hybrid DAG networks becomes possible, and the resultant models can be collected by \textit{Global-DAG} for further involvement.

\smallskip
\noindent\textbf{Roles.} In \textsc{IronForge}, the users can take different roles: \textit{viewer}, \textit{task publisher}, \textit{verifier}, and \textit{contributor}. A user is a participant in the network. Each user can select one or multiple roles to perform specific functional activities (see the left-hand side in Fig.~\ref{fig: model}). Specifically, a \textit{viewer} can directly fetch models from the public resource pool without further actions. The \textit{task publisher} aims to propose new tasks and the proposed tasks are broadcast and await others' contributions. 
In order to reap profits, a user can become a \textit{contributor} to process a training process by selecting, aggregating and training models. He can either start the work on \textit{Global-DAG} or enroll in others' published work from the uncompleted tasks. Also, a \textit{verifier} in the system is to verify existing tasks in the resource pool. He can contribute or verify either one favorable task or multiple tasks in parallel for a higher profit. In short, the four roles cover all potential functional activities within \textsc{IronForge}.

\subsection{Workflow Overview}
Then, we provide an overview of the workflow of \textsc{IronForge}. We focus on the procedures of task establishment and task processing by presenting the interactive steps of a user between the \textit{Task-DAG} and \textit{Global-DAG}. 

\noindent\hangindent 1em  \textbf{\textit{Step-1.}} The user registers a task in the \textit{Global-DAG} network by depositing the committed prize. He obtains a task identifier and then broadcasts the task to the network. We assume that another user has accepted the proposed task prior and worked on the task as a task contributor. 

\noindent\hangindent 1em  \textbf{\textit{Step-2.}} The contributor enters the procedure of training models. He first evaluates several existing models from the pool and selects a series of models for the shortlist.

\noindent\hangindent 1em  \textbf{\textit{Step-3.}} Based on the selected models, the contributor aggregates all the short-listed models and integrates them with local datasets to train the model according to requirements.

\noindent\hangindent 1em  \textbf{\textit{Step-4.}} Once completing the training, the contributor submits the trained model to the \textit{Task-DAG}. Meanwhile, peer contributors may also work on the same task and generate competitively trained models. All these models are propagated within the \textit{Task-DAG} network.

\noindent\hangindent 1em  \textbf{\textit{Step-5.}} The publisher who obtains the trained model terminates the task by marking it with a termination tag. Once selected by users who are conducting the training process in \textit{Global-DAG}, the trained model is deemed to be formally synchronized into \textit{Global-DAG}. 



\smallskip

Notably, a user in the \textit{Global-DAG} network can either contribute to other tasks proposed by peer users, or personally publish a task by himself. All the procedures follow similar steps, as described from \textbf{\textit{Step-2}} to \textbf{\textit{Step-5}}.

\subsection{System Assumptions}

In this section, we list our assumptions on the network, security, and threat models of \textsc{IronForge}. 


\smallskip
\noindent\textbf{Resource assumption.} We do not assume any resource distribution in our work. The resource distribution in the entire network is random. This means different participants, with a high probability, hold different computing resources, including computing power, network bandwidth, memory space, storage capability, and training dataset quality. 
Addressing the system heterogeneity is one of the core contributions in this work, as we weaken the long-existing implicit assumption in previous work~\cite{9524833}: the even resource distribution. \textsc{IronForge} enables any distribution of shares of any type of resources among the participants, making the system practical.

\smallskip
\noindent\textbf{User behavior assumption.} We have two assumptions on user behaviors. First, the participants in the network are \textit{rational}, meaning that they can select an arbitrary task, switch to others, or quit existing tasks for better profits. Second, different participants can focus on different training targets, including both task bundles (one task has dependency on another) and orthogonal tasks (one task is independent of the others). This enables the processing of multiple tasks in parallel, greatly improving the system's overall scalability and performance.

\smallskip
\noindent\textbf{Security assumption.} 
We assume that the honest nodes always conduct honest behaviors, where they obey all the policies during the model selection, model aggregation, model training, task verification, and other operations related to the defense strategies against adversaries. 
The adversaries have the ability to delay the model convergence and lower the model accuracy by leveraging popular FL attacks, including lazy attacks~\cite{9524833}, poisoning attacks~\cite{poisoning}, and backdoor attacks~\cite{backdoor}. 
The adversaries also have the ability to breach the incentive fairness by leveraging model stealing attacks~\cite{stealing-1,stealing-2,stealing-4,stealing-5}, and compromising the privacy of others' training datasets. Adversaries not only can upload their previous models (traditional lazy attacks), but also fake the ownership of others' previous models or fake their own training process to embezzle rewards for their conspirators. These two faking types are defined as stealing attacks and collusion attacks, respectively, and both belong to the context of model stealing attacks in this paper.

\section{Decentralized Federated Learning}\label{sec_DAG}

\begin{table}[!htb]
\caption{Notation Definition }\label{tal: notation}
\label{tab_notation}
\begin{tabular}{p{1.2cm}|p{6.7cm}}
\hline
\textbf{Notation} & \textbf{Definition}   \\
\hline
$U_k$    & The $k$-th user
\\
$B_k$    & The balance of $k$-th user
\\
$D^{\textit{train}}_k$    & The local training dataset of $U_k$
\\
$D^{\textit{test}}_k$    & The local testing dataset of $U_k$
\\
$\beta$ & The number of candidate weights  \\
$\sigma$ & The number of aggregated weights  \\
\hline
\multicolumn{2}{p{3cm}}{\textit{Task-DAG}}\\
\hline
$\textit{Task}_m$    & The $m$-th \textit{Task-DAG} network
\\
$\alpha_m$  & The accuracy target of $\textit{Task}_m$
\\
$\nu_m$ & The committed prize of $\textit{Task}_m$
\\
$U_{m,p}$    & The publisher of $\textit{Task}_m$
\\
$N_{m,g}$    & The genesis node of $\textit{Task}_m$
\\
$T_{m,g}$    & The creation timestamp of $N_{m,g}$ 
\\
$W_{m,g}$    & The initial model weights of $\textit{Task}_m$
\\
$N_{m,k,i}$    & The $i$-th node published by the $k$-th user in $\textit{Task}_m$
\\
$\mathbb{M}_{m,k,i}$ & The source list of $N_{m,k,i}$
\\
$E_{m,k,i}$ & The evaluation result for $N_{m,k,i}$ over $D^{\textit{test}}_k$
\\
$\mathbb{E}_{m,k,i}$ & The evaluation result for $\mathbb{M}_{m,k,i}$ over $D^{\textit{test}}_k$
\\
$W^*_{m,k,i}$ & The model weights aggregated by $\mathbb{M}_{m,k,i}$ before the local training
\\
$W_{m,k,i}$ & The model weights after the local training
\\
$\rho_{m,k,i}$ & The setting for training $W^*_{m,k,i}$ into $W_{m,k,i}$
\\
$T_{m,k,i}$    & The creation timestamp of $N_{m,k,i}$
\\
$N_{m,e}$    & The task-termination node of $\textit{Task}_m$
\\
$T_{m,e}$    & The creation timestamp of $N_{m,e}$
\\
$\Psi_m$ & The prize allocation of $\textit{Task}_m$
\\
$\Phi_m$ & The contest strategy of $\textit{Task}_m$
\\
$\Upsilon_m$ & The penalty strategy for malicious attacks of $\textit{Task}_m$
\\
$\Omega_m$ & The verification committee of $\textit{Task}_m$
\\
\hline
\multicolumn{2}{p{3cm}}{\textit{Global-DAG}}\\
\hline
$N_{g}$    & The genesis node of \textit{Global-DAG}
\\
$N_{k,i}$    & The $i$-th node published by the $k$-th user in \textit{Global-DAG}
\\
$T_{k,i}$    & The creation timestamp of $N_{k,i}$
\\
$S_h$    & The $h$-th settlement node in the settlement sets $\mathbb{S}$
\\
$\lambda_{h}$ & The subtree that is aggregated by $S_h$
\\
$V_{k,k',i}$    & A PoL-challenge raised by $U_{k'}$ for $N_{k,i}$ where $k \neq k'$
\\
$\pi_{k,k',i}$    & The deposit to raise $V_{k,k',i}$
\\
$\epsilon_{\textit{PoL}}$ & The threshold of PoL-verification
\\
$P_{k,k',i}$    & The PoL-response replied by the publisher $U_{k}$ of $N_{k,i}$ for $V_{k,k',i}$ raised by $U_{k'}$ where $k \neq k'$
\\
$R_{k,\hat{k},i}$    & The PoL-result sent from $U_{\hat{k}}$ on $P_{k,k',i}$ where $k \neq \hat{k}$
\\
$\tau_h$    & The timeout for $P_{k,k',i}$ to be published after $V_{k',k,i}$ has been published and confirmed by $S_h$
\\
$\Theta_h$ & The committee elected to conduct consensus for $S_h$
\\

\hline
\end{tabular}
\end{table}

\textsc{IronForge} involves four novel mechanisms, i.e., the release of \textit{Task-DAG} networks, the \textit{decentralized model training}, the \textit{defense strategy}, and the \textit{incentive mechanism}. 
\textsc{IronForge} features two types of network, public \textit{Global-DAG} and task-specific \textit{Task-DAG}.
A training task can be outsourced to communities by releasing a \textit{Task-DAG} network and following four steps including preparation, initialization, monitoring, and finalization. 
The decentralized training processes of \textit{Task-DAG} and \textit{Global-DAG} are specifically defined by the new \textit{decentralized model training} mechanism, in which users aggregate existing models, train the aggregated models and publish the model updates to the network in a decentralized way.
The training processes are guarded by a new \textit{defense strategy} against model stealing attacks in the decentralized setting that has never been considered in existing studies. The crafted \textit{incentive mechanism} assures the state consistency of networks and enables smart-contract-enhanced incentives including both rewards and penalties.
Table~\ref{tab_notation} summarizes the notations.

\subsection{Managing Task-DAG}
Any user can outsource an FL training task by managing a DAG, as shown in Algo.~\ref{alg_manage}. An FL training task can be described with a training target, i.e., the model to be trained, the targeted accuracy, and the testing dataset. To build incentives to motivate distributed workers and deter malicious workers, we design reward, penalty, and verification schemes for \textit{Task-DAG} networks.

\begin{algorithm}[t] 
\footnotesize
\caption{Manage \textit{Task-DAG}} 
\label{alg_manage} 

\nonl {\color{black}\normalsize{$\triangleright$} \textbf{\normalsize{Initialize a training task}}}

\nl $U_{m,p}.$\textbf{Deposit}($\textit{Task}_m, \nu_m$)

$\Omega_m \leftarrow U_{m,p}.$\textbf{VRF}($\textit{Task}_m$)

\nonl\Comment{\textcolor{magenta}{Elect the nominated verification committee to $\textit{Task}_m$}}

$N_{m,p}\leftarrow\{H(W_{m,g}),\ \text{URI}(W_{m,g}),\ H(D_m^{\textit{test}}),\ \alpha_m,$\\
\nonl \qquad\qquad $\nu_m,\ \Phi_m,\ \Upsilon_m,\ \Omega_m,\ T_{m,g}\}$

$N_{m,p}\leftarrow U_{m,p}.$\textbf{Sign}($N_{m,p}$)

$U_{m,p}.$\textbf{Announce}($N_{m,p}$)

\nonl\Comment{\textcolor{magenta}{Broadcast to the network}}

\vspace{0.2cm}
\nonl {\color{black}\normalsize{$\triangleright$} \textbf{\normalsize{Observe the training}}}

\nl\While{True}{
$N_{m,k,i}\leftarrow U_{m,p}.$\textbf{Monitor}($\textit{Task}_m$)

\If{$N_{m,k,i}$ breaches $\Upsilon_m$}{
$U_{m,p}.$\textbf{ApplyPenalty}($\Upsilon_m,\ N_{m,k,i},\ U_k$)
}

\If{$E_{m,k,i}>\alpha_m$}{
$\hat E_{m,k,i}\leftarrow U_{m,p}.$\textbf{Evaluate}$(W_{m,k,i}, D_m^{\textit{test}})$

\If{$\hat E_{m,k,i}>\alpha_m$}{

$\hat N_{m,k,i}\leftarrow N_{m,k,i}$

{\textbf{break}}
}
}
}

\vspace{0.2cm}
\nonl{\color{black}\normalsize{$\triangleright$} \textbf{\normalsize{Finalize the training task}}}

\nl $\Psi_m\leftarrow U_{m,p}.$\textbf{AllocatePrize}($\Phi_m, \hat N_{m,k,i}$)

$N_{m,e}\leftarrow\{\hat N_{m,k,i},\ \hat U_k,\ \text{URI}(D_m^{\textit{test}}),\ \hat E_{m,k,i},\ \Psi_m,\ T_{m,e}\}$
\end{algorithm}

\subsubsection{Preparation}A training task $\textit{Task}_m$ can be described with an initial model, i.e., $W_{m,g}$, and an accuracy target $\alpha_m$ of the model tested on the dataset $D_m^{\textit{test}}$ from the task publisher $U_{m,p}$. To reduce the storage and bandwidth overhead, the model weights can be stored in external infrastructures, e.g., the InterPlanetary File System (IPFS). The Uniform Resource Identifiers (URIs) and hash codes to the weights are embedded in DAG nodes for access and verification. The testing dataset $D_m^{\textit{test}}$ is committed in the genesis node of the \textit{Task-DAG} by embedding the hash code, and is revealed by the end of the training for model verification. This commit-and-reveal design prevents direct access to $D_m^{\textit{test}}$ during the training process and ensures that the final selected model can be publicly verified.

To run a \textit{Task-DAG} with an incentive in a secure way, the task publisher $U_{m,p}$ needs to design a contest strategy $\Phi_m$ to allocate reward $\nu_m$ to contributors and a penalty strategy $\Upsilon_m$ to suppress the flooding of excessive trivial models and other malicious behaviors. 

Some examples of the plug-and-play contest strategy include an egalitarian strategy where the prize is divided equally among contributors along the traversal of the final winner node, or an implementation of ``to each according to his contribution'' whereby the prize is allocated based on the amount of contribution, or striking a balance in-between.
Some examples of the penalty strategy include an implementation of $\frac{\text{S-Index}}{\text{H-Index}}$ (i.e., preventing excessive self-citations)~\cite{s-index} or the occupation ratio along the traversal of the final winner node.
The publisher $U_{m,p}$ also invokes the election to nominate
a set of $U_k$ that constitute a task committee $\Omega_m$ of $\textit{Task}_m$ for evaluating trained models and conducting PoL-verification. The committee is elected via a Verifiable Random Function (VRF)~\cite{vrf} upon the balances $B_k$ of eligible $U_k$.

\subsubsection{Initialization}To initialize $\textit{Task}_m$, the publisher $U_{m,p}$ firstly registers $\textit{Task}_m$ to the task management smart contract $SC_T$ on the \textit{Global-DAG} network by depositing the committed prize $\nu_m$.
Next, $U_{m,p}$ prepares a genesis node $N_{m,g}$ including the commitment and the URI of the initial model to be trained, i.e., $H(W_{m,g})$ and $\text{URI}(W_{m,g})$, the model accuracy target $\alpha_m$, the commitment of the public testing dataset $H(D^{\textit{test}}_m)$, the committed prize $\nu_{m}$, the contest strategy $\Phi_m$, the penalty strategy $\Upsilon_m$, the nominated verification committee $\Omega_m$, and the creation timestamp $T_{m,g}$\footnote{The trustworthiness of the timestamp is guaranteed by trusted timestamping services.}. Then, $U_{m,p}$ can sign the genesis node $N_{m,g}$ and announce the genesis node. 

\subsubsection{Monitoring}Upon these operations, training starts in $\textit{Task}_m$, and the publisher $U_{m,p}$ observes the progress until the model becomes mature enough. 
Any $U_{k}$, who is interested in contributing the computational resources and competing for the prize $\nu_{m}$, continues training models and publishing node $N_{m,k,i}$, i.e., the $i$-th node released by $U_{k}$ in $\textit{Task}_m$, as a worker. If any nodes breaching the penalty strategy $\Upsilon_m$ are found, $U_{m,p}$ issues fines and updates the balance of the publisher of the breaching nodes. Note that the balance change in regard to the penalty has yet to be finalized at this stage.

\subsubsection{Finalization}When the claim of reaching the targeted model accuracy $\alpha_m$ is realized, $U_{m,p}$ evaluates the model over the testing dataset $D_m^{\textit{test}}$. Once the accuracy is surely met by the winner node $\hat N_{m,k,i}$, $U_{m,p}$ executes the contest strategy $\Phi_m$ and obtains the prize allocation $\Psi_m$. Next, $U_{m,p}$ terminates $\textit{Task}_m$ by creating a task-termination node $N_{m,e}$ that points to the winner node $\hat N_{m,k,i}$ and contains the winner's address $\hat U_{k}$, the URI to the testing dataset $\text{URI}(D^{\textit{test}}_m)$, the achieved testing accuracy $\hat E_{m,k,i}$, the prize allocation $\Psi_m$, and the creation timestamp $T_{m,e}$. The task publisher $U_{m,p}$ then signs $N_{m,e}$ and broadcasts $N_{m,e}$ to the \textit{Global-DAG} network. 
The balance change in regard to both the prize allocation $\Psi_m$ and the penalty is subsequently finalized by settlement nodes once the termination node is revealed to the public and is referred by any future model in \textit{Global-DAG}; see details in Section~\ref{subsec: incentive}.

\subsection{Decentralized Model Training}
As an open system, \textsc{IronForge} allows the workers to contribute to the decentralized training in both a $\textit{Task}_m$ and \textit{Global-DAG}. The training process in a \textit{Task-DAG} is given in Algo.~\ref{alg_train}, while the training process in \textit{Global-DAG} shares the same algorithm except that there does not exist a public shared testing dataset $D_m^{\textit{test}}$ that decides the stopping point (i.e., the training accuracy target $\alpha$) in the \textit{Global-DAG} network. \textit{Global-DAG} acts as a public resource pool of diversified models, allowing for free hunting of models that uniquely meet customized training targets upon local testing datasets $D_k^{\textit{test}}$ for each $k$-th user $U_k$.

\begin{algorithm}[t] 
\footnotesize
\caption{Train Models} 
\label{alg_train} 

\For{$U_k$ \textbf{parallelly}}
{

\vspace{0.2cm}
\nonl {\color{black}\normalsize{$\triangleright$} \textbf{\normalsize{Worker registration}}}

\nl $U_{k}.$\textbf{Register}($\textit{Task}_m$)

\While{$\textit{Task}_m$ is ongoing}
{

\vspace{0.2cm}
\nonl {\color{black}\normalsize{$\triangleright$} \textbf{\normalsize{Sync, verify and select nodes}}}

\nl \While{unsynchronized}
{
$N_{m,k',i'}\leftarrow U_{k}.$\textbf{SyncNodes}($\textit{Task}_m$)

$U_{k}.$\textbf{VerifySignature}($N_{m,k',i'}$)

$U_{k}.$\textbf{VerifyRegistration}($U_{k'}$)

$U_{k}.$\textbf{VerifyBalance}($U_{k'}$)

\If{verification passes}{

$U_{k}.$\textbf{Propagate}($N_{m,k',i'}$)


}
}

\For{$\beta \text{ number of }N_{m,k',i'} \in \textit{Task}_m$ \textbf{parallelly}}
{
$E'_{m,k',i'}\leftarrow U_{k}.$\textbf{Evaluate}($W_{m,k',i'},\ D_k^{\textit{test}}$)

$\mathbb{M}_{m,k,i},\ \mathbb{E}_{m,k,i}\leftarrow U_{k}.$\textbf{Select}(\{$N_{m,k',i'},\ E'_{m,k',i'}\}$, $\sigma$)
}

\vspace{0.2cm}
\nonl {\color{black}\normalsize{$\triangleright$} \textbf{\normalsize{Aggregate, train and contribute nodes}}}

\nl $W^*_{m,k,i}\leftarrow U_{k}.$\textbf{Aggregate}($\mathbb{M}_{m,k,i}$)

$W_{m,k,i}\leftarrow U_{k}.\mathcal T_{m,k}(W^*_{m,k,i},\ \rho_{m,k,i},\ D_k^{\textit{train}}$)

$E_{m,k,i}\leftarrow U_{k}.$\textbf{Evaluate}($W_{m,k,i},\ D_k^{\textit{test}}$)

$N_{m,k,i}\leftarrow \{\mathbb{M}_{m,k,i},\ \mathbb{E}_{m,k,i},\ \rho_{m,k,i},$\\
\nonl\quad$H(W_{m,k,i}),\ \text{URI}(W_{m,k,i}),\ E_{m,k,i},\ T_{m,k,i}$\}

$N_{m,k,i}\leftarrow U_{k}.$\textbf{Sign}($N_{m,k,i}$)

$U_{k}.$\textbf{Announce}($N_{m,k,i}$)
}

}
\end{algorithm}

\smallskip
\noindent\textbf{\textit{Task-DAG Training.}}
Any user $U_{k}$, who is interested in competing for the training rewards in a $\textit{Task}_m$, needs to admit the contest and penalty strategies specified in $N_{m,g}$ and registers to the task management smart contract $SC_T$ on the \textit{Global-DAG} network by depositing a certain amount of tokens. This can suppress Sybil attacks, distributed denial-of-service (DDoS) attacks, and other malicious behaviors under the regulation of the penalty strategy $\Upsilon_m$.

While $\textit{Task}_m$ is running, $U_{k}$ can synchronize the view of $\textit{Task}_m$ and obtain the latest nodes $N_{m,k',i'}$  with ($k\neq k' \lor i\neq i') \land (T_{m, k', i'} < T_{m,k,i})$. $U_{k}$ then verifies the signature of the nodes and confirms that the corresponding worker $U_{k'}$ is registered and has enough balance from $SC_T$. Worker $U_{k}$ drops the nodes that do not pass verification and propagates the success nodes to other workers.

After verification, $U_{k}$ randomly evaluates several $N_{m,k',i'}$ over its local testing dataset $D^{\textit{test}}_k$ for the testing accuracy $E'_{m,k',i'}$ until it collects $\beta$ candidated weights (Lines 11-13 of Algo.~\ref{alg_train}). Next, $U_{k}$ picks up the top $\sigma$ models constituting the source list $\mathbb{M}_{m,k,i}$ which are then aggregated into the pre-trained model $W^*_{m,k,i}$ using a weighted aggregation function, as given by 
\begin{equation}
\label{equ_aggregation}
    W^*_{m,k,i}=\sum_{W_p\ \text{in}\ \mathbb M_{m,k,i} \atop E_p\ \text{in}\ \mathbb E_{m,k,i}} \frac{E_p}{\sum_{E_q\in\mathbb E_{m,k,i}} E_q}W_p.
\end{equation}
After that, $U_{k}$ trains the aggregated model $W^*_{m,k,i}$ over its local training dataset $D^{\textit{train}}_k$ with training settings $\rho_{m,k,i}$ for a trained model $W_{m,k,i}$
, as given by 
\begin{equation}
    W_{m,k,i}=\mathcal T_{m,k}(W^*_{m,k,i},\ \rho_{m,k,i},\ D_{k}^{\textit{train}}),
\end{equation}
where $\mathcal T_{m,k}$ is the training function of $U_k$ for $\textit{Task}_m$.

Once the training is done, $U_k$ evaluates the model over its local testing dataset $D_k^{\textit{test}}$ and obtains the testing accuracy $E_{m,k,i}$.
Next, $U_k$ prepares a model update node $N_{m,k,i}$ with the source list $\mathbb{M}_{m,k,i}$, the corresponding accuracy list $\mathbb{E}_{m,k,i}$, the hash code and URI to the trained model, i.e., $H(W_{m,k,i})$ and $\text{URI}(W_{m,k,i})$, the testing accuracy $E_{m,k,i}$, the training settings $\rho_{m,k,i}$, and the creation timestamp $T_{m,k,i}$. Worker $U_k$ then signs $N_{m,k,i}$ and broadcasts the signed node. Note that the training settings $\rho_{m,k,i}$, such as the learning rate and batch size, are embedded in nodes as a record of the training process for any upcoming PoL processes.

\smallskip
\noindent\textbf{\textit{Global-DAG Training.}}
Training in \textit{Global-DAG} also goes through Algo.~\ref{alg_train}, except that there exists neither a public shared testing dataset $D_m^{\textit{test}}$, nor a unique training target that decides the stopping point, hence an indefinitely growing DAG. The testing accuracy $E'_{m,k',i'}$ is also removed in rewarding contributors who upload models to \textit{Global DAG}. $U_k$ receives a reward whenever one of its models gets referred by any other subsequent models in the network, which becomes the default contest strategy for \textit{Global-DAG}. Users conducting training in \textit{Global-DAG} need to be responsible for their own training processes, including preparing their own goals and local testing datasets and hunting for appropriate models across the whole network. Note that, the task-termination nodes of each $\textit{Task}_m$ are also included in \textit{Global-DAG}, which 
enables tasks to be advertised to the wider public and to be further evolvable along with diversified models in \textit{Global-DAG}. 



\subsection{Proof-of-Learning: Defense against Model Stealing Attacks}
\begin{algorithm}[t] 
\footnotesize
\caption{Proof-of-Learning}
\label{algo: pol}
\LinesNumbered 

\nonl {\color{black}\normalsize{$\triangleright$} \textbf{\normalsize{Raise a PoL-challenge}}}

\nl$V_{k,k',i} \leftarrow U_{k'}$.\textbf{Challenge}($N_{k,i} \mid k\neq k' \land
\text{not been challenged}$)

$V_{k,k',i}\leftarrow U_{k'}$.\textbf{Deposit}($\pi_{k,k',i}$)

\nonl\Comment{\textcolor{magenta}{Broadcast to the network}}

$V_{k,k',i} \in \lambda_h$ is subsequently settled by $S_h$.

\nonl\Comment{\textcolor{magenta}{$\tau_h$ countdown starts}}

\vspace{0.2cm}
\nonl{\color{black}\normalsize{$\triangleright$} \textbf{\normalsize{Reply with a PoL-proof}}}

\nl $\widetilde{D}^{\textit{train}}_k \leftarrow$$U_k$.\textbf{Obfuscate}($D^{\textit{train}}_k$)

$P_{k,k',i}\leftarrow$ $U_k$.\textbf{Prove}($V_{k,k',i}$, $\widetilde{D}^{\textit{train}}_k$))

\nonl\Comment{\textcolor{magenta}{Broadcast to the network}}

$P_{k,k',i} \in \lambda_{h+n}$ is subsequently settled by $S_{h+n}$.

\vspace{0.2cm}
\nonl{\color{black}\normalsize{$\triangleright$} \textbf{\normalsize{Verify the PoL-proof}}}

\nl\If{$P_{k,k',i}$ presents before the timeout}
{
\If{$T_{k,k',i}\le T_h+\tau_h$}
{

\For{$U_{\hat{k}} \in \Theta_{h+n+1}$ \textbf{parallelly}}
{
$W^{\textit{replay}}_{k,i}\leftarrow$$U_{\hat{k}}.\textbf{LearningReplay}$($P_{k,k',i}$.($W^*_{k,i}$, $\widetilde{D}^{\textit{train}}_k$, $\rho_{k,i}$))

\If{$\parallel W^{\textit{replay}}_{k,i} - N_{k,i}.W_{k,i}\parallel^2 < \epsilon_{\textit{PoL}}$}
{
$R_{k,\hat{k},i}$ roots for $P_{k,k',i}$
}
}

$\mathcal R_{k,i}\leftarrow$\textbf{Consensus}($R_{k,\hat{k},i} \mid U_{\hat{k}}\in \Theta_{h+n+1}$)

\If{$\mathcal R_{k,i}$ roots for $P_{k,k',i}$}
{
\textbf{emit} \textit{Challenge fails} \Comment{{\textcolor{magenta}{Learning proved}}}

\textbf{goto Finalization}  
}
}
}

\textbf{emit} \textit{Challenge succeeds} {\Comment{\textcolor{magenta}{Learning invalidated}}}



\vspace{0.2cm}
\nonl{\color{black}\normalsize{$\triangleright$} \textbf{\normalsize{Finalization}}}

\nl\If{Challenge fails}
{
$U_{k'}\leftarrow$\textbf{Refund}($\beta\pi_{k,k',i} \mid \beta \in (0,1)$)
}

\Else
{
$U_{k'}\leftarrow$\textbf{Refund}($\pi_{k,k',i}$)

$U_{k'}\leftarrow$\textbf{Penalty}($U_k,\pi_{k,k',i}$)

\textbf{WithdrawReward}($N_{k,i}$)
}





$S_{h+n+1}$ is generated via Algo.~\ref{algo: settle}

\vspace{0.2cm}
\nonl{\color{black}\normalsize{$\triangleright$} \texttt{\normalsize{Notice}}}

\texttt{The \textbf{\textit{Consensus}} is conducted by the nominated verification committee $\Omega_m$ when the PoL is done in a \textit{Task}$_m$.}

\end{algorithm}

We design a privacy-preserving PoL scheme to prove the computing-extensive training work and suppress model stealing attacks~\cite{stealing-1,stealing-2,stealing-4,stealing-5}. The idea of the privacy-preserving PoL is based on the reproducibility of training and the PoL in~\cite{9519402} in which the training process from the same starting point over the same training dataset with the same training settings results in the same trained model or bounded differences among the trained models. In the proposed privacy-preserving PoL, provers can provide obfuscated training dataset and give an estimated bound of model differences. We propose a new dataset obfuscation process to protect the privacy of the training dataset. The proposed privacy-preserving PoL scheme in the \textit{Global-DAG} network is given in Algo.~\ref{algo: pol}. The privacy-preserving PoL scheme for \textit{Task-DAG} networks can be conducted in the same way where the PoL-proof is verified by the nominated task committee $\Omega_m$.

\subsubsection{Challenge}
If a node $N_{k,i}$ has not been challenged before, any worker $N_{k'}$ can raise a PoL challenge against the node as a challenger. $U_{k'}$ needs to deposit a certain amount of tokens $\pi_{k,k',i}$ to the PoL smart contract $SC_{PoL}$ for the challenging node $V_{k,k',i}$. 
Then, $U_{k'}$ signs and broadcasts the challenging node $V_{k,k',i}$ to the network and starts a countdown.

\subsubsection{Response}
The publisher of $N_{k,i}$, i.e., $U_k$, needs to reply to the challenge $V_{k,k',i}$ as a prover within the PoL timeout $\tau$. 
$U_k$ firstly obtains the obfuscated dataset $\widetilde{D}^{\textit{train}}_k$ by applying the noise $\delta_{k,k',i}$ to its local training dataset $D_k^{\textit{train}}$. Next, $U_k$ prepares a PoL proof node $P_{k,k',i}$ with the hash code and URI to the obfuscated dataset, i.e., $H(\widetilde{D}^{\textit{train}}_k)$ and $\text{URI}(\widetilde{D}^{\textit{train}}_k)$. Then, $U_k$ signs $P_{k,k',i}$ and broadcasts it to the network.

\subsubsection{Verification}
At the PoL verification stage, the committee $\Theta_h$ (see details in Section~\ref{subsec: incentive} for the use of $\Theta_h$) verifies $P_{k,k',i}$ in parallel if the timestamp of the prover node is within the timeout $\tau_h$.
To be specific, a verifier $U_{\hat k}$ can fetch the obfuscated dataset $\widetilde{D}^{\textit{\textit{train}}}_k$ with the URI given in $P_{k,k',i}$ and confirm its integrity. Next, $U_{\hat k}$ conducts a training task with the starting model described in $N_{k,i}$, the training settings $\rho_{k,i}$ embedded in $N_{k,i}$, and the training dataset $\widetilde{D}^{\textit{train}}_k$, i.e.,
\begin{equation}
    \begin{split}
    W^*_{k,i}=\sum_{W_p\ \text{in}\ \mathbb M_{k,i} \atop E_p\ \text{in}\ \mathbb E_{k,i}} \frac{E_p}{\sum_{E_q\in\mathbb E_{k,i}} E_q}W_p\\
    W_{k,i}^{\textit{replay}}=\mathcal T_{\hat k}(W^*_{k,i},\ \rho_{k,i},\  \widetilde{D}_{k}^{\textit{train}}).
    \end{split}
\end{equation}
The verifier $U_{\hat k}$ then calculates the Frobenius-Norm (F-Norm) between trained $W_{k,i}^{\textit{replay}}$ and $W_{k,i}$ in $N_{k,i}$ as the PoL result $R_{k,\hat k,i}$ for $P_{k,k',i}$~\cite{self-obfuscation}. If $R_{k,\hat k,i}$ is within the $\epsilon_{\textit{PoL}}$, the verification on $U_{\hat k}$ is success and $R_{k,\hat k,i}$ roots for $P_{k,k',i}$. All versifiers in the committee $\Theta_h$ run consensus algorithms, e.g., Practical Byzantine Fault Tolerance (PBFT)~\cite{castro1999practical}, on PoL results and get the final committee decision $\mathcal R_{k,i}$ as a proving node in the network. Notice that, to prevent the spoofing attacks against the PoL verification and improve the security and robustness~\cite{PoL-attack}, $\epsilon_{\textit{PoL}}$ can be dynamically adjustable from the early stage to the later stage to circumvent the consistent F-norm-based model distance during a PoL spoofing attack. Also, any PoL prover $P_{k,k',i},$ $\forall k,k',i$ could alternatively conduct Verifiable Computation (VC) by showing an additional VC-proof to guarantee the identity of $W_{k,i}^{\textit{replay}}$.

\subsubsection{Clearing}
At the finalization stage, the challenger $U_{k'}$ can only get $\beta\pi_{k,k',i}$ from the challenge deposit, $\beta \in (0,1)$, if the challenge fails (i.e., the learning is proved).
Otherwise, $U_{k'}$ can get a full refund of the challenge deposit and also receive a reward from the penalty on $U_k$. The reward to $U_k$ from $N_{k,i}$ is revoked as well if the learning cannot be proved.

\subsection{Achieving State Consistency: Incentive Basis}
\label{subsec: incentive}
\begin{algorithm}[t]   
\footnotesize
\caption{Node Generation in the Settlement Set}
\label{algo: settle}
\LinesNumbered 

\nonl {\color{black}\normalsize{$\triangleright$} \textbf{\normalsize{Generating a settlement node upon consensus}}}

\nl\While{True}
{
\If{$T \mod \Delta_T = 0$}
{
$\Theta_h \leftarrow$ \textbf{VRF}(\textbf{Balance}($U_k \mid$ $\forall k$))). \Comment{\textcolor{magenta}{Election}}

\While{$\Theta_h$.\textbf{Consensus}($U_k.\textit{view} \mid$ $\forall k \land (U_k \in \Theta_h)$)}
{
\If{consensus is reached}{\textbf{break} and obtain $S_h$}
}


$\mathbb{S}.\textbf{Append}(S_h)$
}
}
\nonl\Comment{\textcolor{magenta}{The latest balance can be found in $\mathbb{S}\lbrack\text{latest}\rbrack$}}

\vspace{0.2cm}
\nonl {\color{black}\normalsize{$\triangleright$} \textbf{\normalsize{Consensus}}}

\nl\For{$U_k$ in $\Theta_h$ \textbf{parallelly}}
{
$\textit{Tips} \leftarrow$ \textbf{Prune}($\lbrace N_{k,i} \mid T_{k,i} > T\rbrace$)

\While{\textbf{Traverse}(start $\leftarrow$\textit{Tips})}
{
\If{(Path-$p$ reaches $N_g$ OR $N_{k,i} \in \lambda_{h-1}$) is \textbf{True}}
{
\textbf{Stop} Path-$p$ 

\If{ALL paths have stopped}
{
\textbf{break} and obtain $\lambda_h$
}
} 
}

$\lambda_h \leftarrow \textbf{Prune}(\textit{Tips})$

\nonl\Comment{\textcolor{magenta}{Obtain the subtree $\lambda_h$ for the creation of $S_h$}}

$\bar{S}_{k,h} \leftarrow$ \textbf{Form}($\lambda_h$.\textit{balance}, $\lambda_h$.\textit{PoL}, $\lbrace N_{m,e} \mid \text{ never been collected by $\mathbb{S}$}\rbrace$)

\nonl\Comment{\textcolor{magenta}{\textit{Tips} are excluded in balance calculation}}

\If{$U_k$ is the leader $U_{\bar k}$ of $\Theta_h$}
{
$\bar{S}_{\textit{leader}} \leftarrow \bar{S}_{k,h}$

\textbf{Broadcast}($\bar{S}_{\textit{leader}}$)
}
\Else
{
\textbf{Verify}($\bar{S}_{\textit{leader}}$, $\bar{S}_{k,h}$)

\If{verification passes}{\textbf{emit} \textit{Consensus is reached}}
\Else{Elect a new $\Theta_h \leftarrow \Theta_h^{'}$ and redo \textbf{Consensus}}
}
}

\end{algorithm}

\textsc{IronForge} features settlement nodes to achieve state consistency for the \textit{Global-DAG} network, enabling smart contracts in DAG and recording consistent account states. The process is shown in Algo.~\ref{algo: settle}.

\textit{Global-DAG} periodically, with the time interval $\Delta_T$, elects the settlement committee $\Theta$ among which consensus is reached to generate settlement nodes.
At the beginning of the $h$-th interval, VRF is used to elect a committee securely $\Theta_h$ for the settlement node $S_h$ and the committee leader $U_{\bar h}$. The probability that any worker $U_k$ is selected for the committee depends on the balance of $U_k$, i.e., $B_k$; see (\ref{eq:election}) below,
\begin{equation}
    \textit{VRF-hash}(\textit{prv}_k, seed) \in \Lambda_k, 
\label{eq:election}
\end{equation}
where $\Lambda_k$ is the area portion occupied by $U_k$ in a hash ring, and $\Lambda_k \propto B_k$.

The committee can then settle the status of \textit{Global-DAG}. To be specific, the leader of the committee $U_{\bar k}$ synchronizes the view of a particular preceding moment of \textit{Global-DAG} via consensus with others, and identifies all the tip nodes in the $h$-th interval, which do not have any successor in the current interval. From each of the tip nodes, $U_{\bar k}$ traverses back according to every preceding node list $\mathbb M_{k,i}$ of $N_{k,i}$ along the path. The search stops and the subtree $\lambda_h$ is obtained when all specific nodes are met, i.e., the first visible node which does belong to the previous subtree $\lambda_{h-1}$ in each path or the genesis node $N_g$. 
Based on all nodes in $\lambda_h$, $U_{\bar k}$ updates the balances of involved workers according to the training contributions, PoL challenges, PoL proofs, and smart contract executions. 
In \textit{Global-DAG}, each valid reference from $N_{k',i}$ awards the owner $U_k$ ($k\neq k'$) of the referred model $N_{k,i}$ with a certain amount of tokens. Next, $U_{\bar k}$ proposes a new settlement node $\bar S_{k,h}$ covering the updated balances and PoL results. The settlement committee $\Theta_h$ verifies and votes $\bar S_{k,h}$. The committee $\Theta_h$ endorses $\bar S_{k,h}$ as $S_h$ if the committee reaches consensus, or elects a new leader otherwise.

\section{Implementation and Evaluation}\label{sec_develop}


In this section, we conduct comparisons between the proposed FL system, \textsc{IronForge}, and other popular frameworks, including GoogleFL~\cite{fl-1}, AsyncFL~\cite{async-fl}, and BlockFL~\cite{8733825}. We experimentally assess \textsc{IronForge} in terms of the model performance and expected amount of rewards that can be earned under a variety of different environment settings, including different aggregation strategies, different sizes of hardware and software resources, and different types and levels of malicious attacks such as lazy attacks~\cite{9524833}, poisoning attacks~\cite{poisoning}, backdoor attacks~\cite{backdoor}, and model stealing attacks~\cite{stealing-1}.

\subsection{Experimental Configurations}
\subsubsection{Hardware settings}
The experiments are conducted on 6 servers listed as follows.

\smallskip
\noindent\textbf{Type-A (\#1-3): }
\begin{itemize}
\item \textbf{CPU. }2 $\times$ Intel(R) Xeon(R) Gold 6230R CPU @ 2.10GHz, 2 $\times$ 52 cores
\item \textbf{GPU. }1 $\times$ Quadro RTX 4000, 1 $\times$ 8GB
\item \textbf{Memory. }528GB
\item \textbf{Bandwidth. }1000Mb/s
\end{itemize}

\smallskip
\noindent\textbf{Type-B (\#4-6): }
\begin{itemize}
\item \textbf{CPU. }2 $\times$ Intel(R) Xeon(R) Gold 6138 CPU @ 2.00GHz, 2 $\times$ 40 cores
\item \textbf{GPU. }8 $\times$ NVIDIA PCIe A100, 8 $\times$ 40GB
\item \textbf{Memory. }250GB
\item \textbf{Bandwidth. }1000Mb/s
\end{itemize}

\smallskip
\subsubsection{Software settings}
We carry out the experiments upon Ubuntu 18.04.6 LTS with Keras 2.7 in Python 3.7.13 and Docker 20.10.12. We use FastDFS as the distributed file system with 15TB storage space for the model weights.

\subsubsection{A new testbed - FLSim}
\begin{figure}[!hbt]
    \centering
    \includegraphics[width=0.9\linewidth]{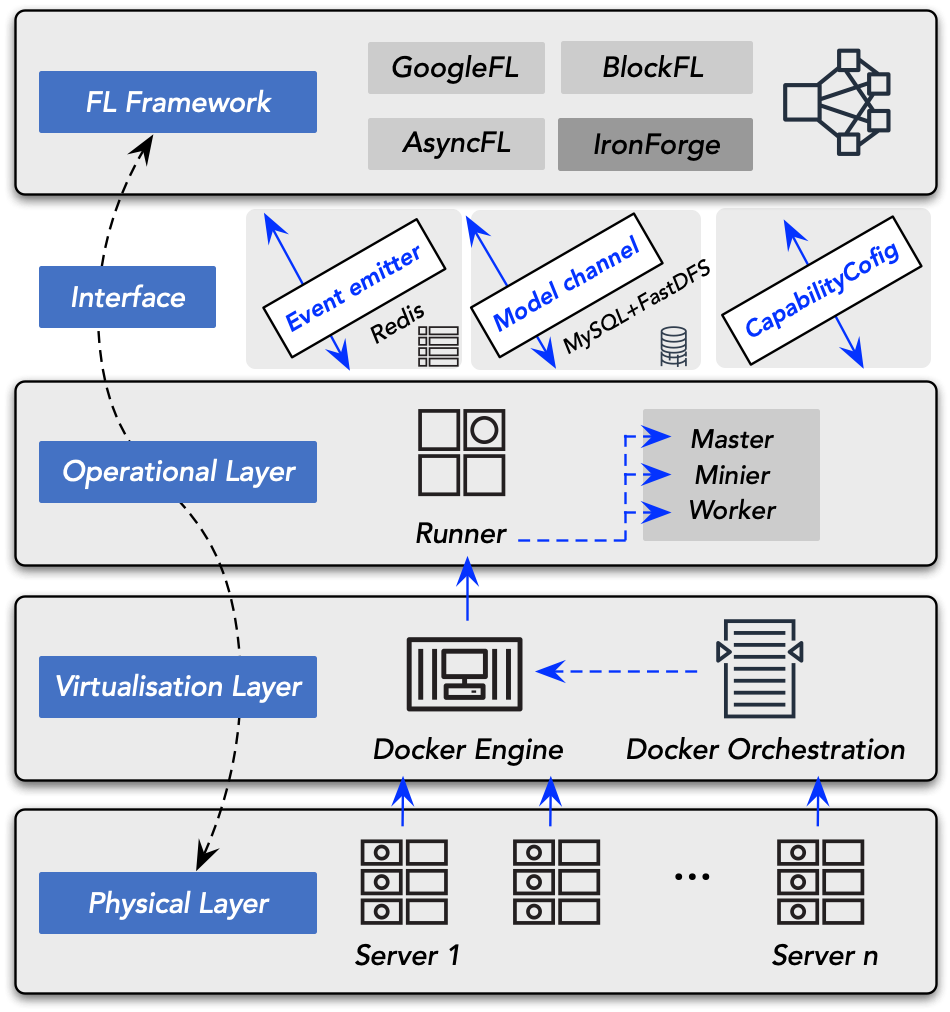}
    \caption{Architecture of the new FL testbed FLSim}
    \label{fig: flsim}
\end{figure}

To benchmark the considered FL frameworks, we build an FL testbed named FLSim as shown in Fig.~\ref{fig: flsim}. 
FLSim is docker-containerized upon our servers (\#1--6).
Choosing different FL frameworks is flexibly plug-and-play in FLSim via three generic interfaces, i.e., the event emitter, model channel, and capability configuration.

\smallskip
\noindent\textbf{Event emitter.}
FLSim is event-driven where all events are delivered through Redis which serves as a message queue. Each runner can receive events in the network in real-time by the subscription function of Redis, and can broadcast corresponding events according to their role.

\smallskip
\noindent\textbf{Model channel.}
Indexing models are done via MySQL where properties such as the URIs of weights are included, while the actual model weights are stored in FastDFS. As a result, the query efficiency can be significantly improved with no need of retaining the large weights unless they are required for evaluation or aggregation.

\smallskip
\noindent\textbf{Capability configuration.}
The tasks are trained on runners deployed on docker clusters. Each docker container represents a runner with different resource settings, such as CPU, memory, and bandwidth. 
Specifications of the containers are craft-specified with strong scalability and flexibility by defining the capability configuration to simulate various scenarios, such as the resource imbalance considered in the experiments. 
Moreover, each runner is categorized into different roles based on which FL framework has been plugged in, e.g., ``workers'' in all considered frameworks, ``masters'' in GoogleFL and AsyncFL, and ``miners'' in BlockFL. 

\subsubsection{Training settings}
\begin{figure}
    \centering
    \includegraphics[width=0.4\textwidth]{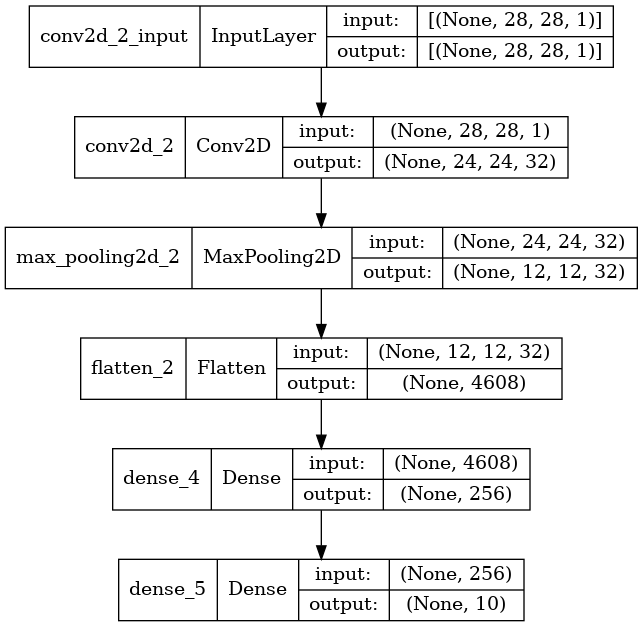}
    \caption{The CNN model is a lightweight version of the model in~\cite{fl-1}. It contains one convolution layer of which the filter size is 32 and the kernel size is 5 $\times$ 5, one 2 $\times$ 2 max-pooling layers, one fully connected layer with 256 units, and ReLu activation. The output is processed by a fully connected layer with 10 units and softmax activation.}
    \label{fig:cnn_model_structure}
\end{figure}

The tuned hyper-parameters of the experiments are summarized in Table~\ref{tal:hyper-parameter}. We perform training over the MNIST with 60,000 data samples and a Convolutional Neural Network (CNN) model illustrated in Fig.~\ref{fig:cnn_model_structure}.

\begin{table}[t]
\renewcommand\arraystretch{1.2}
\caption{Hyper-parameter settings}
\label{tal:hyper-parameter}
\begin{center}
\begin{tabular}{|l|p{4.3cm}|p{1.5cm}|}
\hline
\textbf{Notation} & \textbf{Definition} & \textbf{Value (unit)}\\
\hline
$\mathcal{P}$ & idle probability & 0.1 \\ 
$\mathcal{E}$ & global epoch & 2000 \\ 
$e$ & default local epoch & 5 \\ 
$l$ & learning rate & 0.002 \\
$\eta$ & sampled weights & 30 \\
$\beta$ & default number of candidate weights & 6 \\
$\sigma$ & default number of aggregated weights & 5 \\
$\mathcal B$ & default batch size & 100 \\ 
$\mathcal V$ & validation set size & 100 \\
\hline
\end{tabular}

\end{center}
\end{table}

Totally 60,000 MNIST samples are randomly split into two parts, 48,000 samples are used as the training set and 12,000 samples are used as the testing set. We create non-IID training shards for contributors from the training set to simulate a practical network condition. The training set is divided into two subsets, i.e., 24,000 for each. The samples in the first subset are sorted by labels, and are subsequently distributed into 120 shards, 200 for each shard. Thus, the sample labels in each shard are relatively concentrated. The other half of the samples are randomly selected and distributed into 120 shards, i.e., 200 samples for each shard. Thus, the sample labels in each shard are relatively uniform. Finally, we repeat the sampling operation 120 times, and each time we take a shard from each of the two subsets for merging. We end up obtaining 120 new shards each of which contains 400 samples.

\subsubsection{Environment settings}
To demonstrate the state-of-the-art of \textsc{IronForge}, a comprehensive comparison between \textsc{IronForge} and the other three FL frameworks are conducted in our experiments, i.e., the synchronous GoogleFL, AsyncFL, and BlockFL. We launch 120 runners as workers to train models with a probability of $\mathcal{P}$.
We also define two events that represent receiving two types of intermediate models: 
\begin{itemize}
\item GLOBAL\_MODEL\_UPLOADED\_EVENT (GMUE):
the model acquired by aggregating the uploaded local models prior to training.
\item LOCAL\_MODEL\_UPLOADED\_EVENT (LMUE): 
the model trained by workers with their local datasets
\end{itemize}
These events are broadcast to notify each runner associated with the next step to take. Note that the genesis model of a task is tagged as a global model in order to initiate any selected FL framework. This indicates that emitting GMUE is used to notify the network when the task is published.


\smallskip
\noindent\textbf{Synchronous GoogleFL.} 
One additional runner is launched as the master to aggregate local model weights. For GoogleFL, workers are activated by GMUE and the master is activated by LMUE. The workers, when receiving a GMUE, train on top of a global model downloaded from the master with their own local datasets. The master receives an LMUE when the workers upload the trained models, and subsequently aggregates all collected local models after a timeout, followed by uploading the aggregated result as the global model. The above iterating process continues until the task reached the max iteration threshold $\mathcal{E}$.


\smallskip
\noindent\textbf{AsyncFL.} 
One additional runner is launched as the master to aggregate local models. AsyncFL shares the same procedure as GoogleFL, except that the master, when receiving an LMUE during its idle period, creates a new global model by aggregating the most recent global model and newly-collected local models with an identical weighting factor.


\smallskip
\noindent\textbf{BlockFL.} 
Five additional runners are launched as miners for BlockFL. In each iteration, the miners behave the same way as the master does in GoogleFL or AsyncFL. An additional step is that the miners compute the nonce to finalize the block and compete for the rewards with a synchronized lock being used in Redis to ensure the mining order.

\smallskip
\noindent\textbf{\textsc{IronForge}.} 
No additional runners are launched for \textsc{IronForge}. Each runner acts as a worker and a master at the same time, i.e., it supports both aggregating and training operations, thus receiving both GMU and LMUE during each iteration. 

\subsubsection{Security settings}

We implement five types of contributors: normal contributors, poison contributors, backdoor contributors, and stealing and colluding contributors.

\smallskip
\noindent\textbf{Normal contributors} act honestly and independently across all phases.

\smallskip
\noindent\textbf{Poisoning contributors} aim to undermine the integrity and availability of the global model by crafting local poisoning models~\cite{poisoning}. In this paper, we simulate poison contributors by adopting the label-flipping strategy that fakes labels and then conducting training on the forged datasets. 

\smallskip
\noindent\textbf{Backdoor contributors} aim to fail the global model on targeted tasks, typically by adhering crafted triggers to training samples, conducting training on the amended samples, and then uploading the attack models~\cite{backdoor}. In this paper, backdoor contributors layer $5\times5$ white patches to training samples and change the label of the manipulated samples to a fixed one.

\smallskip
\noindent\textbf{Stealing contributors} aim to gain rewards by stealing model weights trained by others and uploading the plagiarized weights as their own work~\cite{9524833}. In this paper, a stealing contributor $k'$ selects and directly uploads one of the existing weights by simply changing the ownership from $k$ to $k'$.

\smallskip
\noindent\textbf{Colluding contributors} aim to embezzle training rewards for their conspirators by performing honest training processes but claiming source list $\mathbb{M}$ from the conspirators. In this paper, a certain proportion of contributors tamper the source list $\mathbb{M}$ in their uploaded weights, with a certain probability.


\subsection{Results and Evaluations}

Several experiments are conducted from two perspectives, i.e., the performance comparison between \textsc{IronForge} and others with and without attacks, and the fairness comparison between different resource levels in terms of rewards. 

\smallskip
\subsubsection{Performance - with and without attacks}

\begin{figure}[!htb]
    \centering
    \includegraphics[width=0.4\textwidth]{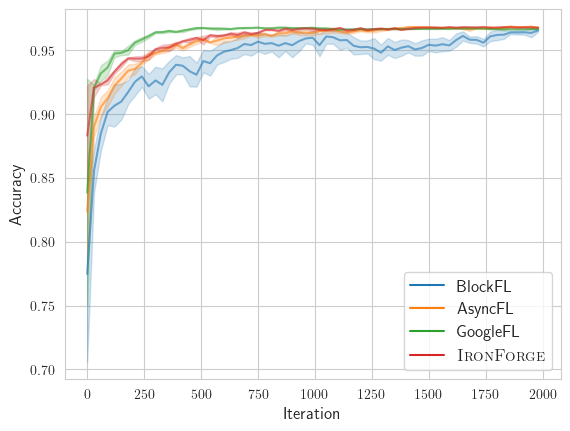}
    \caption{Comparison between \textsc{IronForge} and others in terms of accuracy with no attacks leveraged}
    \label{fig:general-accuracy}
\end{figure}

Fig.~\ref{fig:general-accuracy} shows that the proposed \textsc{IronForge} outperforms AsyncFL and BlockFL with only slightly slower convergence than the baseline GoogleFL across 2,000 iterations with no attacks leveraged. It is worth noting that the red curve increases sharply with as few oscillations as that of the baseline, particularly highlighting the stability of \textsc{IronForge}.

\begin{figure*}[t]
    \centering
    \subfigure[Stealing attacks]{
    \begin{minipage}[t]{0.31\textwidth}
    \centering
    \includegraphics[width=2.4in]{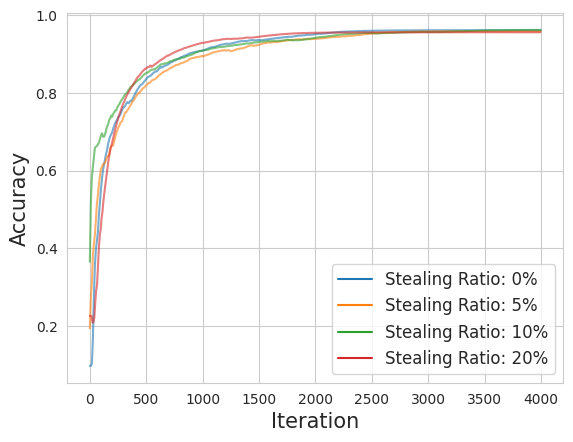}
    \end{minipage}
    \label{fig:exp_3_stealing}
    }
    \subfigure[Poisoning attacks]{
    \begin{minipage}[t]{0.31\textwidth}
    \centering
    \includegraphics[width=2.4in]{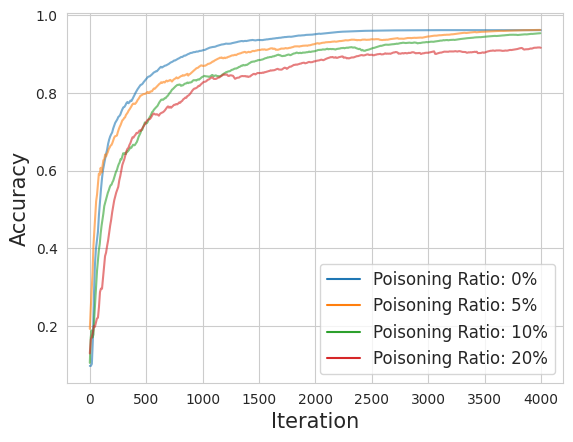}
    \end{minipage}
    \label{fig:exp_3_poisoning}
    }
    \subfigure[Backdoor attacks]{
    \begin{minipage}[t]{0.31\textwidth}
    \centering
    \includegraphics[width=2.4in]{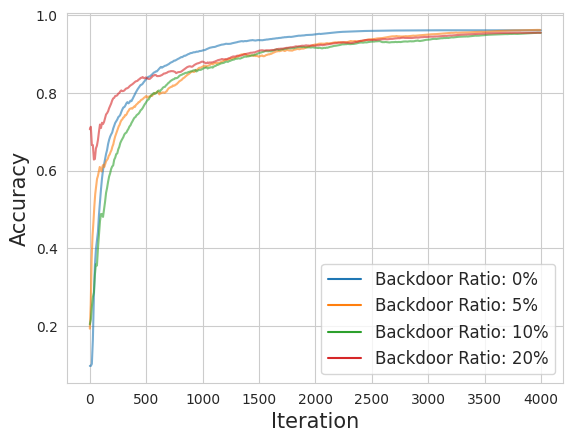}
    \end{minipage}
    \label{fig:exp_3_backdoor}
    }
\caption{Comparison between different levels of stealing attacks, poisoning attacks, and backdoor attacks applied to \textsc{IronForge} in terms of accuracy}
\label{fig:attack-accuracy-1}
\end{figure*}

\begin{figure*}[t]
    \centering
    \subfigure[Poisoning attacks with a poisoning ratio of 20\%]{
    \begin{minipage}[t]{0.31\textwidth}
    \centering
    \includegraphics[width=2.4in]{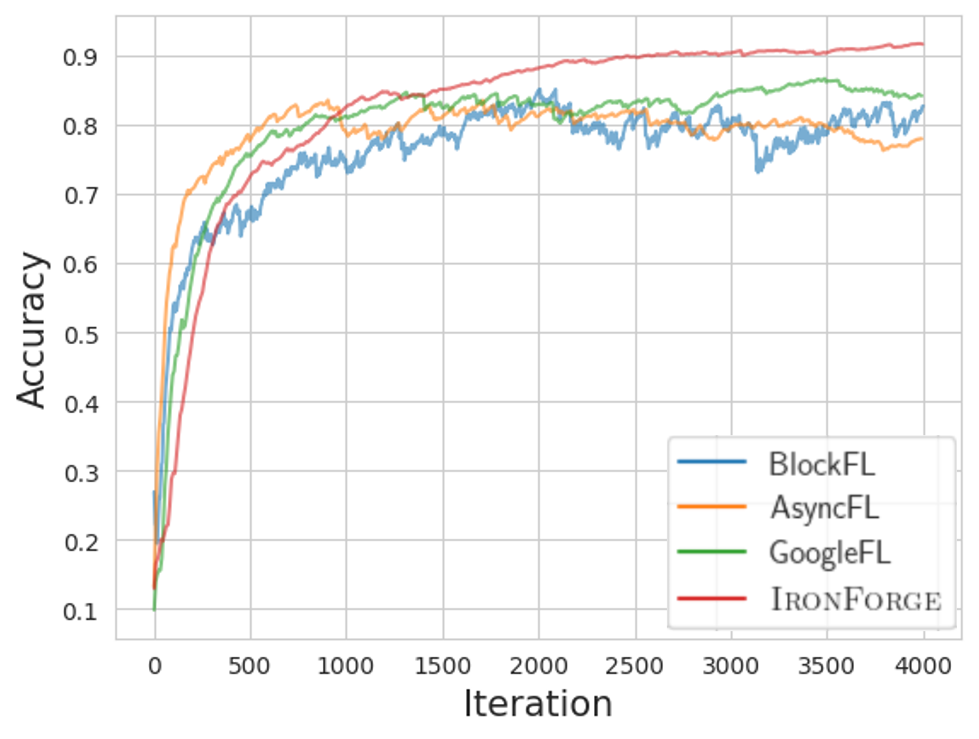}
    \end{minipage}
    \label{fig:exp_4_poisoning}
    }
    \subfigure[Backdoor attacks with a poisoning ratio of 20\%]{
    \begin{minipage}[t]{0.31\textwidth}
    \centering
    \includegraphics[width=2.4in]{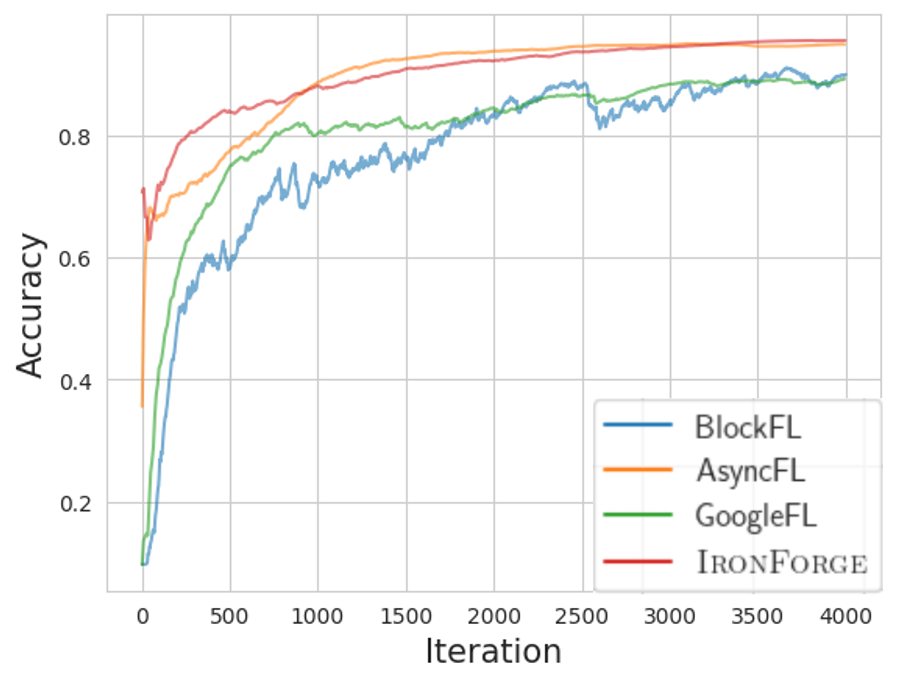}
    \end{minipage}
    \label{fig:exp_4_backdoor}
    }
    \subfigure[Stealing and collusion attacks]{
    \begin{minipage}[t]{0.31\textwidth}
    \centering
    \includegraphics[width=2.4in]{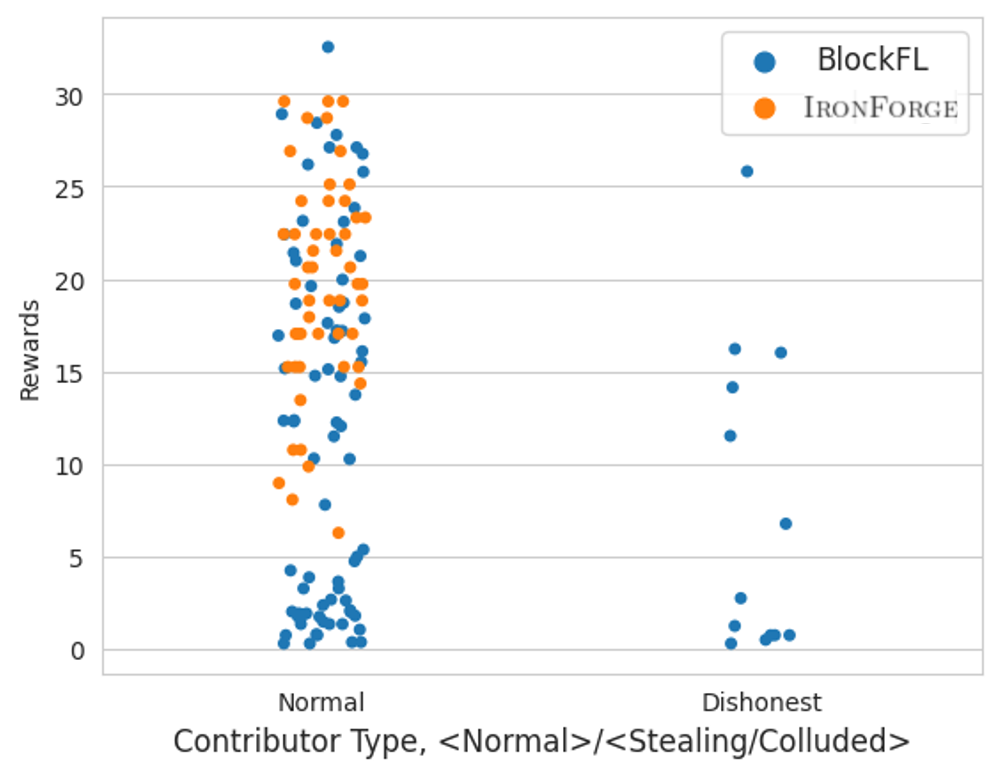}
    \end{minipage}
    \label{fig:exp_5_pol}
    }
\caption{Comparison between \textsc{IronForge} and others in terms of accuracy or rewards with a certain level of poisoning attacks, backdoor attacks, and stealing and colluding attacks}
\label{fig:attack-accuracy-2}
\end{figure*}

\begin{figure*}[t]
    \centering
    \subfigure[Accuracy difference between aggregation sizes]{
    \begin{minipage}[t]{0.31\textwidth}
    \centering
    \includegraphics[width=2.4in]{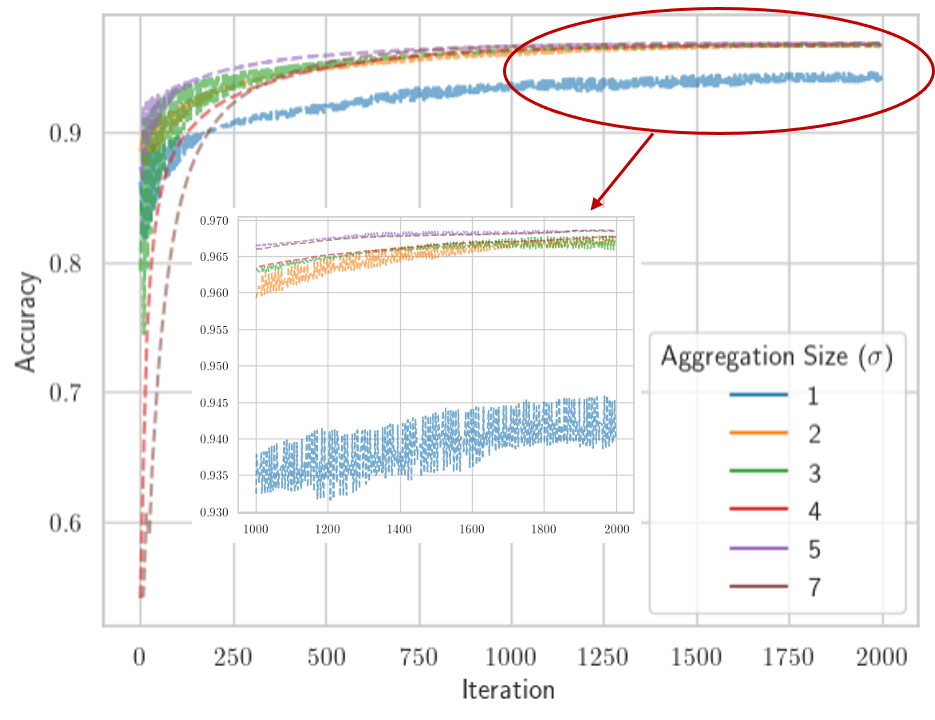}
    \end{minipage}
    \label{fig:exp_6_mn_1}
    }
    \subfigure[Accuracy difference between candidate sizes]{
    \begin{minipage}[t]{0.31\textwidth}
    \centering
    \includegraphics[width=2.3in]{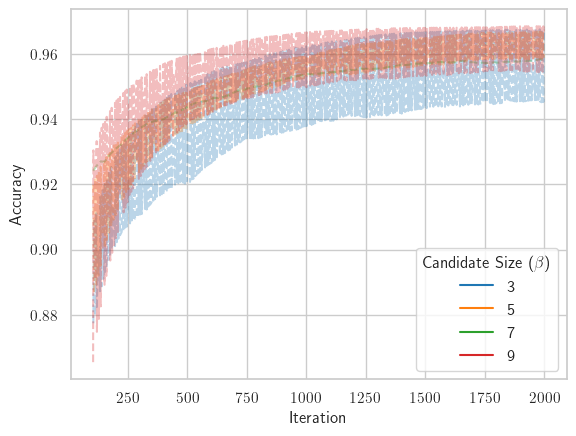}
    \end{minipage}
    \label{fig:exp_6_mn_2}
    }
    \subfigure[Time difference between aggregation strategies]{
    \begin{minipage}[t]{0.31\textwidth}
    \centering
    \includegraphics[width=2.4in]{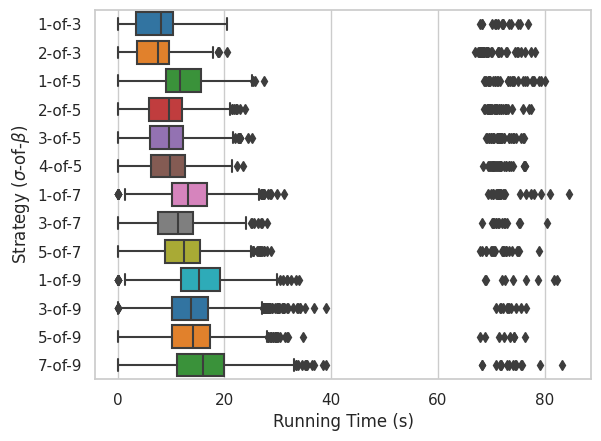}
    \end{minipage}
    \label{fig:exp_6_mn_3}
    }
\caption{Comparison between different combination of the aggregation strategies ($\sigma$-of-$\beta$) applied to \textsc{IronForge} in terms of accuracy and execution time}
\label{fig:aggregation-strategy}
\end{figure*}

The performance comparison with different levels of attack behaviors being applied to \textsc{IronForge} is shown in Fig.~\ref{fig:attack-accuracy-1}. It is realized that \textsc{IronForge} is resistant to the stealing attack the most, followed by the resistance to the backdoor attacks and poisoning attacks. It is worth noting that the performance of a stealing ratio of 20\% can be as good as that of others. This is because the native validation process in \textsc{IronForge} can capture and eliminate the plagiarized models which are reused or whose ownerships are fake. The remaining 80\% of models are still sufficient for contributors to aggregate and train by offering strong diversity of the non-IID data samples.

On the other hand, an evident degradation of the accuracy, around 5\%, is shown in both poisoning (cf. Fig.~\ref{fig:exp_3_poisoning}) and backdoor (cf. Fig.~\ref{fig:exp_3_backdoor}) contexts. Nevertheless, it can be found from Fig.~\ref{fig:exp_4_poisoning} and~\textcolor{violet}{\ref{fig:exp_4_backdoor}} that \textsc{IronForge} outperforms all the other FL frameworks with either 20\% ratio of poisoning attackers or 20\% ratio of backdoor attackers. This significantly highlights the superiority of \textsc{IronForge} in terms of its strong resistance to malicious model updating.
Fig.~\ref{fig:exp_5_pol} highlights the resistance to stealing attacks and collusion attacks of \textsc{IronForge}. Note that only BlockFL is considered in the comparison as GoogleFL and AsyncFL do not support incentives natively. It is found that, by using the native validation process in \textsc{IronForge}, including the PoL verification, none of the dishonest contributors who leverage either the stealing attack or collusion attack can gain rewards. This prevents the malicious contributors from faking the ownership of or directly using the existing models, or embezzling the rewards for their conspirators by claiming a falsified source list.

\begin{figure*}[t]
    \centering
    \subfigure[Reward difference between different CPU cores]{
    \begin{minipage}[t]{0.31\textwidth}
    \centering
    \includegraphics[width=2.4in]{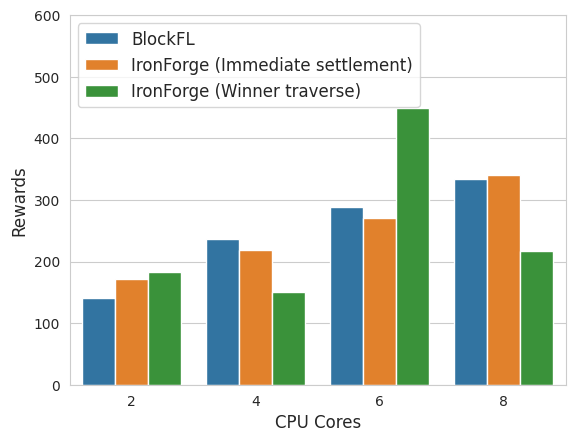}
    \end{minipage}
    \label{fig:exp_2_cpu}
    }
    \subfigure[Reward difference between memory capacity]{
    \begin{minipage}[t]{0.31\textwidth}
    \centering
    \includegraphics[width=2.4in]{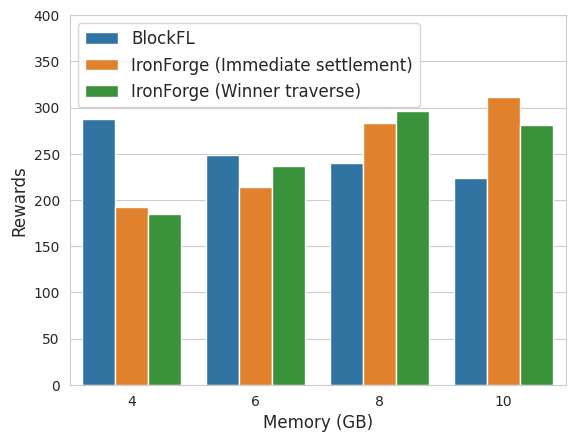}
    \end{minipage}
    \label{fig:exp_2_memory}
    }
    \subfigure[Reward difference between bandwidths]{
    \begin{minipage}[t]{0.31\textwidth}
    \centering
    \includegraphics[width=2.4in]{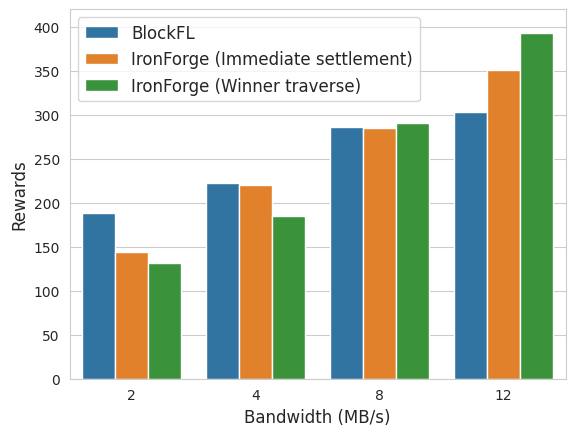}
    \end{minipage}
    \label{fig:exp_2_bandwidth}
    }
\caption{Comparison between different levels of the CPU core, memory capacity, and bandwidth in terms of rewards}
\label{fig:fairness-1}
\end{figure*}

\begin{figure}
    \centering
    \includegraphics[width=0.37\textwidth]{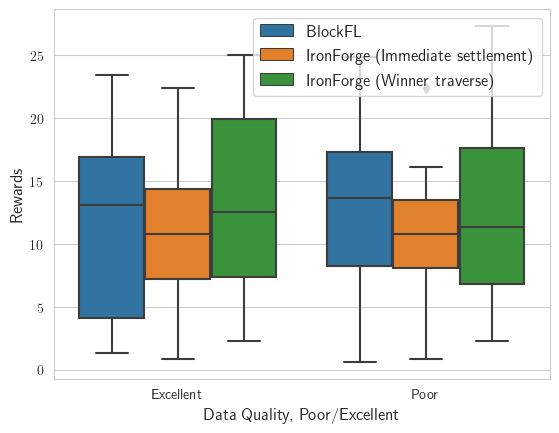}
    \caption{Comparison between \textsc{IronForge} and others in terms of rewards under the influence of the data quality issue}
    \label{fig:exp_2_data_quality}
\end{figure}

Fig.~\ref{fig:exp_6_mn_1} shows the accuracy ranges of adjusting the candidate size ($\beta$) for different levels of aggregation sizes ($\sigma$), e.g., the blue band representing the accuracy ranged from 1-of-2 to 1-of-8 with $\sigma=1$ and $\beta\in[2,8]$. The results reveal that increasing the aggregation size $\sigma$ stabilizes the performance in the beginning stage, and allows for an increasingly higher convergence point. The effect of increasing the candidate size for a certain aggregation size
becomes gradually weakened as the aggregation size increases, as you can see that the width of each band turns more and more narrow. 
The effect of increasing the aggregation size also becomes weakened, as you can see from Fig.~\ref{fig:exp_6_mn_1} that the performance of 5-of-6 is as good as that of 7-of-8.
The same insight can also be shed by observing the performance comparison of adjusting the aggregation sizes ($\sigma$) for different levels of candidate size ($\beta$) is shown in Fig.~\ref{fig:exp_6_mn_2}, e.g., the blue band representing the accuracy ranged from 1-of-3 to 2-of-3. The bottom line of the red band for $\beta=9$ performs as poorly as the 2-of-3 strategy does, while the top line of the red band performs the best among the kinds. It can thus be concluded that knowing that no attacks are leveraged, aggregating more weights is more beneficial for obtaining high accuracy than merely aggregating a low number of weights from more candidate weights.

Fig.~\ref{fig:exp_6_mn_3} shows the running time of different aggregation strategies all the way from downloading the weights to uploading a new model. There is a stronger effect on the running time when adjusting the candidate size ($\beta$) than adjusting the aggregation size ($\sigma$). An implication can be realized based on this observation that the bandwidth could be the bottleneck in \textsc{IronForge}.
We design dedicated experiments that learn this phenomenon, as explained in the following Section~\ref{fairness}.

\smallskip
\subsubsection{Fairness - earn rewards}\label{fairness}

The fairness is investigated in terms of the difference in rewards between different levels of hardware specifications. The bottleneck of gaining more rewards in \textsc{IronForge} can also be realized. We select two different contest strategies, i.e., Immediate settlement and Winner traverse. The immediate-settlement is the strategy used in \textit{Global-DAG} by default, rewarding every model every time it gets referred by others. The winner-traverse could be one of the main options used in a \textit{Task-DAG}, rewarding every model that exists in the traversal path all the way from the winner node to the genesis node (excluded).

According to Fig.~\ref{fig:fairness-1}, A monotonic increase of the rewards can be realized for the immediate-settlement with increasing CPU cores, memory capacity, and bandwidth, and for the winner-traverse only with increasing bandwidth (the winner-traverse appears to have similar characteristics to BlockFL which also fails to offer a pure monotonic increase of rewards in all specifications). This is because the winner-traverse strategy could include moderate models being aggregated during each iteration in the winner-traversal path while the winner could be highly random when the competition is intense and the convergence is near. Therefore, many models with high performance could be excluded by the unique winner traversal path at the end. On the contrary, the immediate-settlement allows every model not to be missed so long as a valid reference relationship is confirmed. Nevertheless, the result difference between these two strategies does not tell the superiority of fairness. Different requirements may lead to different principles of fairness. Rooting for an egalitarian strategy, or ``to each according to his contribution'', or striking a balance in between is a flexible option that the proposed \textsc{IronForge} offers back to users without a harsh setting.

On the other hand, the monotonic increase in the bandwidth comparison for both strategies, as shown in Fig.~\ref{fig:exp_2_bandwidth}, highlights the bandwidth being the most critical effect for earning rewards in \textsc{IronForge}. That is to say, relatively poorer users being more active in uploading models to the network with higher frequency can help them be more likely to share the rewards, rather than spending much time on a strongly performant model.

Fig.~\ref{fig:exp_2_data_quality} learns the effect of data quality upon the rewards by adding a random perturbation to 50\% of the training samples of half of the users with a mean of 0 and a standard deviation of 1. We define the affected nodes as ``Poor'' nodes, while ``Excellent'' nodes own normal data samples only. 
The result shows that the poor nodes that use the immediate-settlement strategy enjoy a narrower range of rewards compared to that of the winner-traverse strategy and BlockFL. This highlights that poor nodes earning rewards via the immediate-settlement in \textsc{IronForge} can be more stable and predictable than BlockFL and the winner-traverse, and can be less affected by unexpected data degradation or network noise in unreliable channels. Excellent nodes have more opportunities to earn higher rewards than poor nodes while the median value and the minima of rewards remain as high as that of poor nodes. 
This reflects the fairness between excellent and poor nodes, i.e., offering stable rewards to the poor while the excellent are given chances to make a great fortune.

\section{Discussion and Analysis}\label{sec_analysis}


This analysis focuses on the security of \textsc{IronForge}. Every role in the system is involved in the attack model except that the timestamp in \textsc{IronForge} is considered synchronous via external trustworthy servers.

Adversaries target to break the state consistency in \textit{Global-DAG} so that operations such as incentive and consensus fail to be executed. 
The adversaries also target to leverage the model stealing attack in order to:
\begin{itemize}
\item forge the ``amount of work'' by simply stealing others' models with no more effort being put into the training;
\item collude with attackers by creating a model referring to models from colluded attackers.
\end{itemize}
In addition, the adversaries can target on breaching the dataset privacy during the dataset sharing in a PoL process. At the same time, the adversaries can also unbalance the competition by abusing others' datasets to enrich local resources.

\smallskip
\noindent\textbf{State consistency:} The state consistency is guaranteed over a sufficiently long period $\Delta_T$ as long as the seeds being used in each VRF process are secure and the lower bounds of faulty tolerance of consensus protocol (e.g., 33\% for PBFT) are satisfied in the VRF-elected committees. We consider the time gap between two settlement nodes $\Delta_{S_h, S_{h-1}}$ is sufficiently large in \textsc{IronForge} to expect that each user who gets registered for committee election has an identical ``view'' of \textit{Global-DAG} starting from $T_{h-1}$ to $T_{h-1}+\Delta_T$. This prevents the consensus process in the committee from being trapped into an indefinite disagreement due to the network asynchrony. On the other hand, an unbiased and unpredictable random seed is crucial for a fair VRF process where (\ref{eq:election}) cannot be manipulated. This can be achieved by implementing existing randomness generators such as RANDAO~\cite{randao} or RandHound-VRF~\cite{omniledger}.

\smallskip
\noindent\textbf{Model stealing attack:} Offering the proposed incentive mechanism in a decentralized FL attracts attackers to leverage model stealing attacks, by either stealing the model ownership or faking the training processes. Attackers can, with no effort on local training, steal others' models and fake ownership with ease. Attackers can alternatively fake the source lists upon an honest local training process so that their accomplices, who are instead placed in the source list, can reap profits against other honest users. These two types of model stealing attacks are used for misleading those who wish to ensure the necessary training overhead and the efforts in the source list. They are particularly useful when attackers intend to steal the rewards and share them with their accomplices. \textsc{IronForge} enables PoL-challenge where users can choose to challenge a model via idle resources, and the model owner requires to provide the valid PoL-proof in time for a public verification during the consensus process. By the committee replaying parts of the training from scratch and reaching the consensus, the ``amount of work'' and the source list $\mathbb{M}$ can be explicitly determined.

\smallskip
\noindent\textbf{Dataset privacy and model melting:} This security metric is an implementation of our work~\cite{self-obfuscation}. Dataset obfuscation helps to preserve dataset privacy when datasets require to be publicly shared for PoL-challenge. Experimental results in~\cite{self-obfuscation} show that an obfuscated dataset satisfies PoL verification without sacrificing the privacy level while being able to decrease the model utility against the abuse of collecting provers' obfuscated datasets, namely, model melting.
 In addition, applying training over different data samples or using non-IID noise significantly can reduce the risks of privacy decline when a sufficient number of challenges against the same model owner are deliberately raised by attackers.

\section{Conclusion}
\label{sec_con}
This paper proposed \textsc{IronForge}, a new generation of FL framework constructed by DAG-based structure, which for the first time eliminates the need for the central coordinator to solve the issues of network asynchrony and the excessive reliance on the central coordinator while at the same time enabling an open and fair incentive mechanism to encourage more participants, particularly in networks where training resources are unevenly distributed. 
Experimental results based on a newly developed testbed FLSim along with the security analysis
highlight the superiority of \textsc{IronForge} over the existing prevalent FL frameworks under various specifications regarding performance, fairness, and security. 
To our knowledge, this is the first paper proposing a secure and fully decentralized FL framework that can be applied in open networks with realistic network and training settings.

\normalem
\bibliographystyle{IEEEtran}
\bibliography{bare_jrnl}

\end{document}